\documentclass[pdflatex,sn-basic]{sn-jnl}

\usepackage{graphicx}%
\usepackage{multirow}%
\usepackage{amsmath,amssymb,amsfonts}%
\usepackage{amsthm}%
\usepackage{mathrsfs}%
\usepackage[title]{appendix}%
\usepackage{xcolor}%
\usepackage{textcomp}%
\usepackage{manyfoot}%
\usepackage{booktabs}%

\usepackage{algorithm}      
\usepackage{algpseudocode}

\usepackage{listings}%

\usepackage[small]{caption}
\usepackage{graphicx}
\usepackage{amsmath}
\usepackage{amsthm}

\usepackage[switch]{lineno}
\usepackage{tabularray}
\usepackage{diagbox}
\usepackage{amsthm}
\usepackage{multirow}
\usepackage{subcaption}
\usepackage{placeins}
\usepackage{tabularx}
\usepackage{wrapfig}  
\usepackage{float}      
\usepackage{caption}    

\theoremstyle{thmstyleone}%
\newtheorem{theorem}{Theorem}
\theoremstyle{thmstyletwo}%
\newtheorem{example}{Example}%

\theoremstyle{thmstylethree}%
\newtheorem{definition}{Definition}%

\begin{document}


\title[Article Title]{Ranked Set Sampling-Based Multilayer Perceptron: Improving Generalization via Variance-Based Bounds}


\author[1, 2]{ \sur{Feijiang Li}}\email{fjli@sxu.edu.cn}

\author[1]{ \sur{Liuya Zhang}}\email{202322408058@email.sxu.edu.cn}

\author[1, 2]{ \sur{Jieting Wang}}\email{jtwang@sxu.edu.cn}

\author[1, 2]{ \sur{Tao Yan}}\email{hongyanyutian@sxu.edu.cn}

\author*[1, 2]{ \sur{Yuhua Qian}}\email{jinchengqyh@126.com}

\affil[1]{\orgdiv{Institute of Big Data Science and Industry, Shanxi University, Taiyuan, China}}

\affil[2]{\orgdiv{Key Laboratory of Evolutionary Science Intelligence of Shanxi Province, Taiyuan, China}}




\abstract{

Multilayer perceptron (MLP), one of the most fundamental neural networks, is extensively utilized for classification and regression tasks. In this paper, we establish a new generalization error bound, which reveals how the variance of empirical loss influences the generalization ability of the learning model. Inspired by this learning bound, we advocate to reduce the variance of empirical loss to enhance the ability of MLP. As is well-known, bagging is a popular ensemble method to realize variance reduction. However, bagging produces the base training data sets by the Simple Random Sampling (SRS) method, which exhibits a high degree of randomness. To handle this issue, we introduce an ordered structure in the training data set by Rank Set Sampling (RSS) to further reduce the variance of loss and develop a RSS-MLP method. Theoretical results show that the variance of empirical exponential loss and the logistic loss estimated by RSS are smaller than those estimated by SRS, respectively. To validate the performance of RSS-MLP, we conduct comparison experiments on twelve benchmark data sets in terms of the two convex loss functions under two fusion methods. Extensive experimental results and analysis illustrate the effectiveness and rationality of the propose method. (The code will be made publicly available.)
}



\maketitle

\section{Introduction}
Multilayer Perceptron (MLP) is the fundamental building block of deep learning and also the most basic form of neural networks. It consists of multiple layers of neurons, each layer fully connected to the subsequent one~\cite{DeepLearning}. 
MLP has been extensively applied to various classification and regression problems, including time series analysis area~\cite{Chen2023TSMixerAA,Zeng2022AreTE}, phoneme recognition~\cite{Sivaram2011Sparse,Sivaram2011Multilayer}, image classification~\cite{VIT,LIU2022108829}, domain adaption~\cite{10506994} and so on.

Despite its widespread use, MLP faces challenges in terms of over-fitting and generalization performance.
Consequently, research has focused on improving MLP's predictive accuracy by incorporating techniques like regularization~\cite{Clara2024,Reshetova2024}, ensemble methods~\cite{TruongThanh2025}, and modified training method~\cite{Liu2023}. 
This study concentrates on classification tasks, with a particular emphasis on exploring innovative ensemble methods to enhance MLP's generalization capability, aiming to provide novel solutions for optimizing the performance of this classic neural network architecture.

The evaluation of the generalization ability of machine learning models is one of the core issues in statistical learning theory. 
The generalization ability of existing learning models is often analyzed through frameworks such as VC dimension ~\cite{Mohri2018VC}, Rademacher complexity~\cite{Bartlett2001Rademacher, wang2020learning}, margin distribution~\cite{GAO20131,Goodman1979MeasuresOA}, stability analysis \cite{10.1162/153244302760200704, li2019clustering}, and PAC-Bayes theory \cite{McAllester2013}. 
While these approaches are theoretically significant, they predominantly focus on the complexity of the overall hypothesis space while neglecting the crucial influence of data distribution on generalization. 
Traditional generalization bounds tend to be overly loose, failing to accurately reflect true model performance. 
Model variance, a critical metric for assessing prediction stability under specific data distributions, is directly linked to generalization capabilities. We extend existing theory by deriving a widely applicable error bound that establishes the connection between empirical loss variance and generalization. Our findings demonstrate that decreasing variance can substantially enhance model performance.

As a classical technique for reducing model variance, Bagging ensemble constructs a set of base classifiers via Simple Random Sampling (SRS)~\cite{Dietterich2000Ensemble}, effectively mitigating variability across different training sets and significantly improving the stability and accuracy of the model~\cite{Ganaie2021EnsembleDL,Van2012EnsembleMF,Opitz1995GeneratingAA}.
However, in scenarios where data contain substantial noise or exhibit uneven distribution, traditional SRS sampling encounters dual limitations. First, when data include significant noise or uneven distribution, the high similarity among the base classifier training datasets may not sufficiently reduce variance; Furthermore, repeated sampling decreases diversity between base classifiers, potentially leading to suboptimal overall model performance~\cite{Wang2023RSS}.

To address the inherent shortcomings of SRS, Ranking Set Sampling (RSS) has introduced a paradigm shift in sampling methodology by incorporating a ranking mechanism~\cite{Mart2006Pruning}. 
Theoretical studies demonstrate that RSS ensures more uniform coverage of the sample space through hierarchical sorting while preserving the randomness inherent in SRS. Taking advantage of the importance of sample order to avoid large variance issues in random sampling  ~\cite{Stokes1988CharacterizationOA, Bai2003OnTT, Chen1999DensityEU}, RSS introduces a ranking mechanism that dynamically adjusts the proportion of category representation, effectively alleviating the impact of imbalanced distributions.

Based on the above viewpoint, to enhance the performance of MLP by reducing the variance of loss estimation, we perform ranked sampling on the dataset. 
On the sampled subsets, we construct multiple MLPs and aggregate their predictions to obtain the final result.
To theoretically analyze the performance of the method, we consider the deviation between the empirical loss and the expected loss as the variable to be bounded. Using Lipschitz continuity, we obtain an upper bound for the deviation with respect to sample variations. Subsequently, we employ McDiarmid's inequality to establish an upper bound for the deviation concerning its expected deviation. Finally, Jensen's inequality is applied to derive an upper bound for the expected deviation in terms of variance.
This paper further proves that for MLP, ensemble strategies based on RSS achieve superior variance reduction effects compared to purely random sampling.
In summary, the main contributions are as follows:
\begin{itemize}
    \item A new  general framework for analyzing generalization performance and variance is developed, which elucidates the impact of loss variance under empirical distributions on the performance of classification algorithms.
    \item The expectation and variance of empirical loss on RSS samples and that on SRS samples are theoretically compared, which explains why RSS performs well.
    \item A multilayer perceptron bagging ensemble model is proposed using Ranked Set Sampling that reduces base classifiers' empirical loss variance. 
    Experiments confirm its effectiveness across multiple settings.
\end{itemize}

The proofs are put in the appendix section.

\section{Related Work}
The main contents involved in this paper are generalization performance bound and ranked set sampling method, and we will review from these two aspects.
\subsection{Generalization Bound}
The learnable theory begins in the 1960s and developed into a relatively flourish theoretical system in the 1990s~\cite{Valiant1984theory}.
The generalization ability of existing learning models is often analyzed through frameworks such as VC dimension ~\cite{Valiant1984theory,Vapni2000TheNO}, Rademacher complexity~\cite{Dembczy2017Consistency,Wang2023Generalization}, and margin distribution~\cite{schapire1998boosting,Gao2013On}.
Besides, Fisher information is utilized to describe the generalization performance of learning models~\cite{Wang2023RSS}.
For kernel based loss functions, the generalization bounds based on Rademacher Chaos Complexities and some other new proposed complexity measures are given~\cite{6795617,CLEMENCON}.
For deep learning model, spectrally-norm-based and sub-Gaussian norm-based are given~\cite{Neyshabur2017,Neyshabur15}.
For convex loss function, the relationship between zero-one risk and substitute risk is quantified~\cite{Bartlett2006ConvexityCA,2004Statistical}. Some convergence bounds based on convex theory are also proposed to reveal the effective of convex loss optimization method~\cite{Cheng2024DeepEC,JMLR:v23:20-1135}.
In this paper, we consider establishing a bound in terms of the variance of empirical convex loss function.
\subsection{Ranked Set Sampling}
The RSS method originated from the estimation of average forage grass and fodder production~\cite{Halls1966Trialof}. While the cost of measuring an object can be quite high, sorting a small number of objects is relatively straightforward. In such cases, the RSS method can enhance estimation accuracy and reduce sampling costs. An improved version of RSS method and its application in parameter estimation are presented, and the effectiveness of ordered sampling with different distributions is demonstrated~\cite{Bouza2014Review,Barabesi2001efficiency}.
Given the function form of probability density distribution, parameter estimation solution based on RSS sample are presented~\cite{he2020maximum,yang2020efficiency}.
In estimating the Shannon entropy and Rayleigh entropy, the performance of samples obtained through RSS is superior to that derived from simple random sampling (SRS)~\cite{Jafari2014On}.
Classification method based on finite mixture models is proposed under RSS sample~\cite{hatefi2014estimation}.
In this paper, we consider using RSS samples to improve the performance of MLP.

\section{A Variance-Based Generalization Bound}

Given a hypothesis function space $\mathcal{F}$,
the task of classification is to learn a function $f(X)\in \mathcal{F}$ mapping from the feature space $X\in\mathcal{X}\subseteq\mathcal{R}^{d}$ to the discrete label space $Y\in\mathcal{Y}$, and the learnt function is desired to have a good classification ability with minimal zero-one risk $\mathit{R}^{*} = \inf \mathit{R}(f) $, where $\mathit{R}(f)$ is the expected zero-one risk
$\mathit{R}(f)=\mathbb{E}_{X,Y}\mathbb{I}\{Y\neq f(\textit{\textbf{X}})\}$, $\mathbb{I}\{{\cdot}\}$ is the indicator function. Since the indicator function is a step-shaped function that is difficult to optimize, it is commonly replaced by a convex loss $\phi$, which can be effectively optimized by the gradient method.
Correspondingly, the expected convex risk is 
$\mathit{R}_{\phi}(f)=\mathbb{E}_{X,Y} \phi(Y, f(\textit{\textbf{X}}))$.

Usually, the underlying probability distribution of $\mathcal{X}\times\mathcal{Y}$ is unknown.
We merely have a set of independently empirical data drawn from it, denoted as
$D=\{(\textbf{\textit{x}}_1,y_1),...,(\textbf{\textit{x}}_N,y_N)\}$, where $N$ is the number of object.
Through the training data set, classifiers can be obtained by optimizing the empirical convex risk functions:
$\hat{\mathit{R}}_{\phi}(f)=\frac{1}{N} \sum_{i=1}^{N} \phi(y_i,f(\textit{\textbf{x}}_i))$ .

To measure the generalization ability of classifiers, many works have been proposed to investigate whether the excess risk $\mathit{R}(f)-\mathit{R}^{*}$ can convergent to zero as the sample size increases.
\cite{Bartlett2006ConvexityCA} presents a novelty view to study the relation between excess risks in terms of zero-one loss $\mathit{R}(f) - \mathit{R}^{*}$ and convex losses $\mathit{R}_{\phi}(f) - \mathit{R}_{\phi}^{*}$. Inspired by this work, we further give the bound of excess convex risk $\mathit{R}_{\phi}(f) - \mathit{R}_{\phi}^{*}$ in Theorem \ref{th1}
to show what factors and how the factors influence the performance of classifiers, which can be used to inspire some new learning algorithm. 

\begin{theorem}\label{th1}
Let $\mathcal{F}: \mathcal{X}^d \rightarrow \mathbb{R}$ denote the hypothesis space containing the hypothesis function $f(\textit{\textbf{x}})$ with a bounded norm. Assume that $\phi(\textit{\textbf{x}})$ is a loss function that is classified
and Lipschitz continuous with a constant $L$. Then, with probability at least $1 - \delta$ (where $\delta > 0$), we have:
{\fontsize{8pt}{8pt}\selectfont
 \begin{align}
        \mathit{R}(f) - \mathit{R}^{*}
          \le \psi^{-1} \bigg (2\underset{f\in \mathcal{F}}{\sup}\sqrt{ \frac{ 2\left(\mathrm{L}\max_\textit{\textbf{x}}\left\|f(\textit{\textbf{x}}) \right\| \right)^2}{N}
        \ln{\frac{1}{\delta }}}
        +2\underset{f\in \mathcal{F}}{\sup}\sqrt{ \mathbb{V} \left ( \hat{\mathit{R}}_{\phi}(f) \right ) }  +\inf_{f\in \mathcal{F}} \mathit{R}_{\phi}(f) - \mathit{R}_{\phi}^{*} \bigg),
        \label{Dt}
    \end{align}
}
   
 where $ \mathbb{V} (\cdot)$ is the variance, $\psi$ is a nondecreasing function, $ \mathit{R}_{\phi}(f) = \mathbb{E}_{\phi}(Yf(X)) $ is the expected convex risk, and
$\mathit{R}_{\phi}^{*}= \inf \mathit{R}_{\phi}(f)$ is the global optimal convex risk.
\end{theorem}

For exponential loss function, the inverse function
$\psi^{-1} (\theta)$ takes the following form: $\psi^{-1} (\theta) = \sqrt{1-\left (1-\theta\right ) ^2}$.
The form of $\psi$ for some classification-calibrated loss functions $\phi$ can refer to Section 2.4 in~\cite{Bartlett2006ConvexityCA}.

Theorem \ref{th1} states that if the sample size tends to infinity,
then the empirical optimal hypothesis function $\hat{f}= \mathop{\arg\min}{_{f}} \hat{\mathit{R}}_{\phi}(f) $ can achieve the optimal zero-one loss ($\mathit{R}(\hat{f}) \rightarrow \mathit{R}^{*}$) with an order of $\mathcal{O}(\sqrt{\frac{1}{N}})$. In addition,
if the variance of the empirical convex risk is small, the generalization ability of classifier is stronger.

\section{Ranked Set Sampling Based MLP Method}

Guided by Theorem \ref{th1}, we proposed Ranked Set Sampling based MLP Method (RSS-MLP) to reduce the variance of the empirical loss. We first introduce ranked set sampling (RSS), including its processes and theoretical comparison with random sampling (SRS). Subsequently, we introduce the RSS-MLP.

\subsection{Ranked Set Sampling}

RSS is a probability sampling method for estimating population parameters. It ensures samples various levels of the population distribution, reducing sampling error, and then improving estimation efficiency. The primary objective of RSS is to obtain a sample set with superior distributional representativeness. 

\subsubsection{The Process of RSS}



For the training set with $N$ objects $D =\{(\textbf{\textit{x}}_1, y_1), (\textbf{\textit{x}}_2, y_2), \ldots, (\textbf{\textit{x}}_N, y_N)\}$, the RSS forms a training set $Dt = \{ Dt_{1},Dt_{2},...,Dt_{\lfloor N/K \rfloor}\}$, where $K$ is an integer less than N and $Dt_i$ is the set of samples formed by the $i$th ranked sampling.

To form $Dt_i$, we randomly select $K^2$ objects from $D$ without replacement. These $K^2$ objects are randomly divided into $K$ groups, each containing $K$ objects. Next, we apply a sorting function $s(\textit{\textbf{x}})$ to arrange the objects within each group. After sorting, we select one object from each group to form $Dt_{i}$. Specifically, we choose the smallest object from group 1, the second smallest from group 2, and so on, select $K$th smallest from group $K$. Consequently, the set $Dt_ {i}$ consists of $ Dt_{i} =\{ X_{[1]i},y_{[1]i} X_{[2]i}, y_{[2]i} , ... , X_{[K]i},y_{[K]i}\}$.

\subsubsection{Theoretical Results on Convex Loss Functions}

Theorem \ref{th1} establishes the generalization bound with respect to the variance of empirical convex loss. Theoretically, we compare the expectation and variance of empirical loss estimated by SRS sample and RSS sample: $\hat{R }_{\phi,SRS}(\textit{f})$ and $\hat{R }_{\phi,RSS}(\textit{f})$.

The following theorems indicate that the expectations of $\hat{R }_{\phi,SRS}(\textit{f})$ and $\hat{R }_{\phi,RSS}(\textit{f})$ are equal, while the variance of $\hat{R }_{\phi,RSS}(\textit{f})$ is smaller than of $\hat{R }_{\phi,SRS}(\textit{f})$.

\begin{theorem}(\textbf{Equal Expectation})
\label{L_Expect}
For the additive convex loss function $\phi$ and an independent identically distribution sample, the expectations of empirical risks based on the SRS sample and RSS sample are equal:
$\mathbb{E} \hat{R} _{\phi,RSS}(f) = \mathbb {E}\hat{R}_{\phi,SRS}(f)$.
\end{theorem}
\begin{theorem}(\textbf{Smaller Variance})
For the additive loss function $\phi$ and an independent identically distribution sample, the variances of the empirical risk based on the SRS and RSS samples satisfy:
$\mathbb{V} \hat{R}_{\phi,RSS}(f) \le \mathbb{V} \hat{R }_{\phi,SRS}(f)$.
 \label{L_Var}
\end{theorem}
The inequality in Theorem \ref{L_Var} mainly due to the Jensen's inequality for convex functions.
From Theorem \ref{L_Var}, we know that the empirical risk estimated by the RSS sample has a smaller variance. Combined with
Theorem \ref{th1}, we can conclude that using the empirical loss estimated by the RSS sample to guide the algorithm may generate a model that has a smaller generalization error and a strong generalization ability.

\subsubsection{Theoretical Results on Exponential Loss and Logistic Loss}

In this section, we will present specific variance difference results obtained for RSS and SRS for two convex loss functions in the sense of binary classification.

\begin{example}(\textbf{Exponential Loss (ExpL) })
For the exponential loss function $\phi (\alpha)= e^{- \alpha}$ with the property of convexity and additive property, we have,
{\fontsize{8pt}{8pt}\selectfont
    \begin{equation}
    \label{Exp_Var}
        \begin{split}
          \mathbb{V} \hat{R }_{\phi,RSS}(f) 
          &=  \mathbb{V} \hat{R }_{\phi,SRS}(f)+\left(\frac{e-e^{-1}}{2}\right)^{2}\frac{1}{Km}
           \left [ \left [\mathbb{E}(yf(\textit{\textbf{x}}))\right ]^{2}-\frac{1}{K} \sum_{r=1}^{K} \left [\mathbb{E}(y_{[r]}f(\textit{\textbf{x}}_{[r]}))\right ]^{2} \right ].
        \end{split}
    \end{equation}
}
\end{example}

\begin{example}(\textbf{Logistic Loss (LogL)})
For the logistic loss function $\phi (\alpha)= log(1+e^{- \alpha})$ with the property of convexity and additive property, we have,
    \begin{equation}
    \label{Log_Var}
        \begin{split}
          \mathbb{V} \hat{R }_{\phi,RSS}(f) 
          &=  \mathbb{V} \hat{R }_{\phi,SRS}(f)
          +\frac{1}{4Km}\left [\left(\mathbb{E}(yf(\textit{\textbf{x}}))\right)^{2}-\frac{1}{K} \sum_{r=1}^{K} \left(\mathbb{E}(y_{[r]}f(\textit{\textbf{x}}_{[r]}))\right)^{2}\right ].
        \end{split}
    \end{equation}
\end{example}
Example 1 and Example 2 give the specific differences between variances. From the differences, we can
observe that the variance gap between RSS and SRS is in an order of $\mathcal{O}(\frac{1}{N})$ (for $m=\lfloor N/K \rfloor$ ), where $N$ is the number of object. The proofs of the Example 1 and Example 2 are given in Appendix.

\subsection{The Process of RSS-MLP}

\begin{wrapfigure}[15]{r}{0.5\textwidth} 
\begin{minipage}[t]{0.47\textwidth}
    \centering
    \vspace{-0.5em} 
    \includegraphics[width=\linewidth]{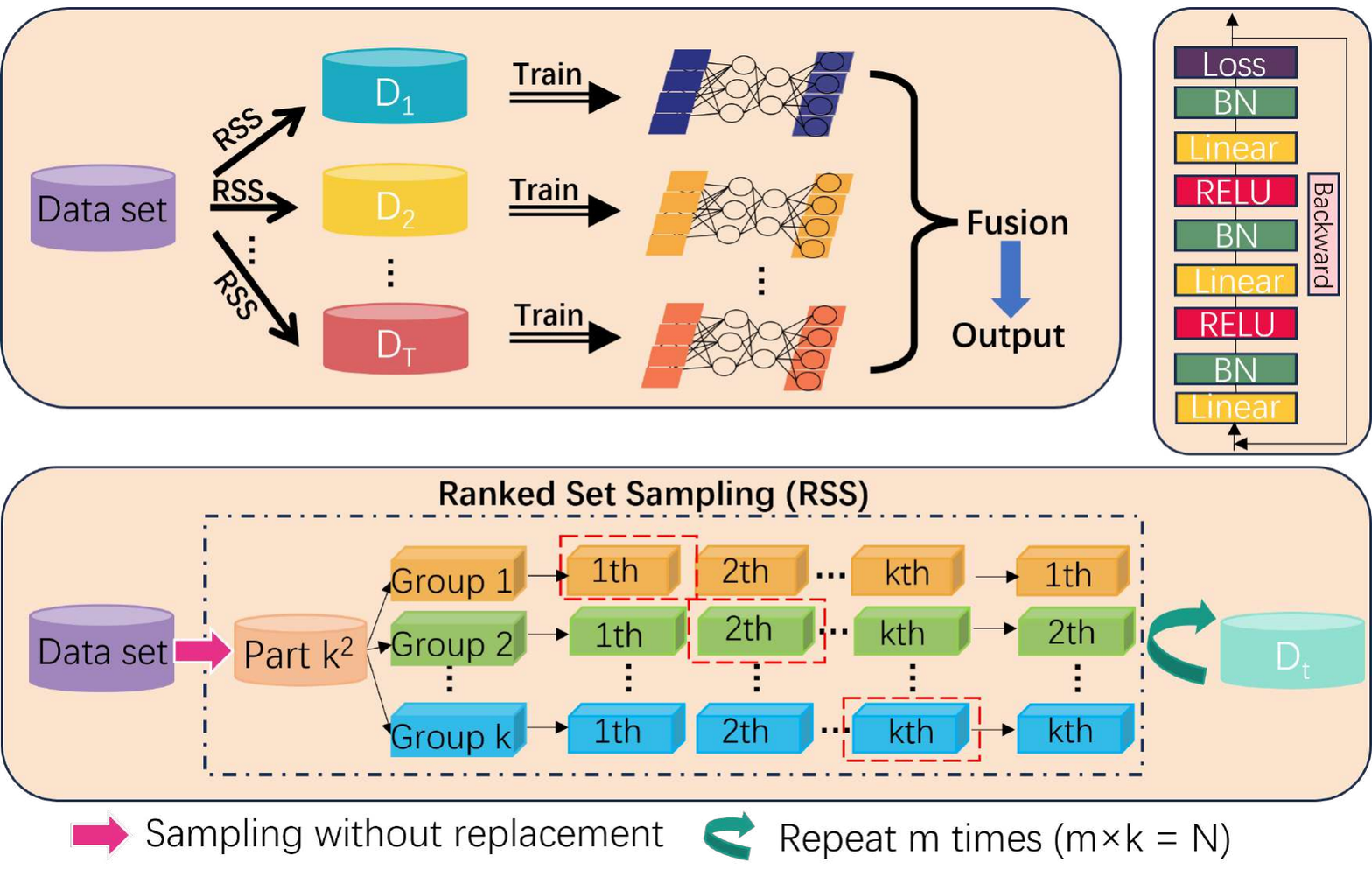}
    \captionof{figure}{The Framework of RSS-MLP}
    \label{framework MLP-Bagging}
\vspace{-20pt}
\end{minipage}%
\end{wrapfigure}

The main structure of the RSS-MLP algorithm is illustrated in Figure \ref{framework MLP-Bagging}, comprising three primary components: ranked set sampling, base classifier training, and classifier ensemble. The processes are shown in Algorithm \ref{MLP-Bagging}.

\begin{wrapfigure}{r}{0.5\textwidth} 
\vspace{-75pt}
  \begin{minipage}{\linewidth}
    \begin{algorithm}[H]
        \caption{RSS-MLP}
        \label{MLP-Bagging}
        \begin{algorithmic}[1]
            \State \textbf{INPUT:} The number of base classifiers $T$, the sampling parameter $K$, the rank function $s$, and the training set $D=\{(\textbf{\textit{x}}_1, y_1), \ldots, (\textbf{\textit{x}}_N, y_N)\}$.
            \State \textbf{OUTPUT:} The set of base classifiers $F = \{f_1,f_2,\dots,f_T\}$.
            \State \textbf{Process:}
            \For{$t = 1$ to $T$}
                \State Perform bootstrap sampling on $D$ using the RSS method to get $D_t$.
                \State Train a three-layer fully connected weak classifier $f_t(D_t)$.
            \EndFor
        \end{algorithmic}
    \end{algorithm}
  \end{minipage}
\vspace{-10pt}
\end{wrapfigure}


The RSS method constructs multiple data sets for the training of base classifiers by multiple resamples of the original dataset. Each resampled subset introduces necessary variability through random sampling while preserving the representative structure of the original data, thereby effectively enhancing the diversity among the base classifiers. 


It should be noted that existing studies indicate that the sorting rules of samples have a relatively minor impact on the final performance of RSS \cite{Bai2003OnTT}. In this paper, we propose using the Spearman rank correlation coefficient as the basis for ranking the data features. Subsequently, the features are ranked in descending order, prioritizing highly discriminative features for constructing the subspace. This non-parametric correlation measure effectively captures the monotonic relationship between features and target variables. 
This ensures the samples with a higher discriminative ability.

The ensemble methodology integrates several weak classifiers to form a robust model aimed at improving classification performance. Each multi-layer perceptron serves as a base classifier, trained on the resampled datasets using either an exponential $exp(-\alpha)$ or logistic loss function $log(1+exp(-\alpha))$. Subsequently, the output results of various MLPs are integrated through voting mechanisms or averaging strategies. This process aims to effectively consolidate the predictive outcomes of resampled instances.

\section{Experimental Analyses}
In this section, we evaluate the performance of the proposed RSS-MLP method from seven aspects, including Comparison Experiments, Ensemble Number Analysis, Violin Diagram, Time Analysis, Comparison with Variance Reducing Techniques, Parameter Analysis, and Significance Analysis.

\subsection{Experimental Settings}

\textbf{Datasets.} The baseline includes twelve datasets from the UCI repository and image data CIFAR-10. For image feature extraction, we adopt the ViT-B/16 architecture with default hyperparameters as provided in \cite{VIT}. Table \ref{dataset} provides a detailed description of the datasets. 

\begin{wraptable}{r}{0.5\textwidth} 
\begin{minipage}[t]{0.5\textwidth}
    \centering
   \vspace{-20pt}
    \captionof{table}{Description of the artificial data sets}
    \setlength{\tabcolsep}{1pt}
    \renewcommand{\arraystretch}{0.8}
    \small
    \begin{tabular}{ccccc}
    \hline
            ID & Name & Objects & Features & Class \\
            \hline
            1  & Cardiotocography   & 2126  & 21  & 10  \\
            2  & Energy             & 768   & 8   & 3   \\
            3  & Abalone            & 4177  & 8   & 3   \\
            4  & Blood              & 478   & 4   & 2   \\
            5  & Magic              & 19020 & 10  & 2   \\
            6  & Twonorm            & 7400  & 20  & 2   \\
            7  & Optical            & 3823  & 62  & 10  \\
            8  & Contrac            & 1473  & 9   & 3   \\
            9  & Waveform           & 5000  & 21  & 3   \\
            10 & Wine-Quality       & 1599  & 11  & 6   \\
            11 & Letter             & 20000 & 16  & 26  \\
            12 & Oocytes-Merluccius & 1022  & 25  & 3   \\
            13 & CIFAR-10           & 60000 & 32×32×3  & 10   \\
    \hline
    \end{tabular}
    \label{dataset}
\vspace{-10pt}
\end{minipage}
\end{wraptable}

\textbf{Evaluation Method and Metrics.} For each dataset, we conduct a random partition into training and testing sets with a ratio of $7$:$3$. This partition is executed 30 times to obtain a comparison of average performance. We employ accuracy and $F_{1}$-score to evaluate the performance of each method. The larger values indicate a better classification performance.

\textbf{Comparison Methods.} 
The bagging ensemble method with the SRS sampling method is chosen as the comparison method. 
For each ensemble method, two fusion strategies and two convex loss functions are implemented. Thus, each algorithm corresponds to four variants. The two fusion methods are voting rules and mean rules.
The two convex loss function algorithms are ExpL and LogL.

\textbf{Implementation Details.} The weak classifier used is a network with three fully connected layers, whose structure is shown in Figure \ref{framework MLP-Bagging}.
Each layer of the network consists of a linear layer, a batch normalization layer, and a ReLU activation layer. The output dimensions of the three layers are set to 256, 128 and the number of classes. The number of base classifiers $T$ is set to $T=51$. 
The $K$ in Eq. (\ref{Dt}) is set as $\lfloor\sqrt{N}\rfloor$, which is the maximum possible number of classes for N samples~\cite{LI2017389, li2025k, li2023fuzzy, li2021got}. 
The batch size number is set to $32$. 

\subsection{Experimental Results}

\textbf{Comparison Experimental Results.} Table \ref{table_exp_acc}-Table \ref{table_log_f1} show the accuracy values and $F_1$-scores of the two sampling methods with exponential loss and logistic loss. 
The situations where the average performance of RSS-MLP exceeds that of SRS-MLP are displayed in bold. If RSS-MLP significantly outperforms SRS-MLP based on a one-sided t-test at a 5\% significance level, the results are marked with an asterisk symbol. The experimental results demonstrate that RSS-MLP uniformly performs better across different datasets. In addition, it can be found that the mean integration method demonstrates superior performance compared to the voting method on almost all datasets and under different settings.

\begin{table*}[!ht]
\vspace{-10pt}
\caption{Comparison of SRS and RSS accuracy under exponential loss for different \(T\) values, with analysis of results via voting and mean methods under different settings}
\label{table_exp_acc}
\tiny
\setlength{\tabcolsep}{1pt} 
\begin{tabular*}{\linewidth}{@{\extracolsep{\fill}}l*{12}{c}}
\hline
\multicolumn{1}{l}{\multirow{2}{*}{Id}}& 
\multicolumn{6}{c}{exp\_vote\_Accuracy}& 
\multicolumn{6}{c}{exp\_mean\_Accuracy} \\
\cmidrule(lr){2-7} \cmidrule(lr){8-13}
& \multicolumn{2}{c}{$T=31$}
& \multicolumn{2}{c}{$T=41$}
& \multicolumn{2}{c}{$T=51$}
& \multicolumn{2}{c}{$T=31$}
& \multicolumn{2}{c}{$T=41$}
& \multicolumn{2}{c}{$T=51$} \\
\cmidrule(lr){2-3} \cmidrule(lr){4-5} \cmidrule(lr){6-7} 
\cmidrule(lr){8-9} \cmidrule(lr){10-11} \cmidrule(lr){12-13}
& SRS & RSS & SRS & RSS & SRS & RSS & SRS & RSS & SRS & RSS & SRS & RSS \\
\hline
1  & 0.8994$\ast$& \textbf{0.9130}  & 0.9001$\ast$& \textbf{0.9125}   & 0.8999$\ast$& \textbf{0.9125}    & 0.8973$\ast$& \textbf{0.9086}   & 0.8981$\ast$& \textbf{0.9084} & 0.8979$\ast$& \textbf{0.9082} \\

2  & 0.8508$\ast$& \textbf{0.8772}  & 0.8508$\ast$& \textbf{0.8756}   & 0.8522$\ast$& \textbf{0.8768}    & 0.8553$\ast$& \textbf{0.8798}   & 0.8545$\ast$& \textbf{0.8788} & 0.8547$\ast$& \textbf{0.8780} \\

3  & 0.6515$\ast$& \textbf{0.6674}  & 0.6518$\ast$& \textbf{0.6678}   & 0.6519$\ast$& \textbf{0.6681}    & 0.6520$\ast$& \textbf{0.6631}   & 0.6517$\ast$& \textbf{0.6632} & 0.6516$\ast$& \textbf{0.6641} \\

4  & 0.7610$\ast$& \textbf{0.7873}  & 0.7600$\ast$& \textbf{0.7883}   & 0.7600$\ast$& \textbf{0.7897}    & 0.7545$\ast$& \textbf{0.7817}   & 0.7545$\ast$& \textbf{0.7813} & 0.7564$\ast$& \textbf{0.7817} \\

5  & 0.8640$\ast$& \textbf{0.8682}  & 0.8642$\ast$& \textbf{0.8686}   & 0.8643$\ast$& \textbf{0.8685}    & 0.8637$\ast$& \textbf{0.8689}   & 0.8639$\ast$& \textbf{0.8689} & 0.8642$\ast$& \textbf{0.8690} \\

6  & 0.9719$\ast$& \textbf{0.9758}  & 0.9720$\ast$& \textbf{0.9757}   & 0.9721$\ast$& \textbf{0.9756}    & 0.9713$\ast$& \textbf{0.9753}   & 0.9714$\ast$& \textbf{0.9755} & 0.9716$\ast$& \textbf{0.9755} \\

7  & 0.9737$\ast$& \textbf{0.9797}  & 0.9734$\ast$& \textbf{0.9798}   & 0.9738$\ast$& \textbf{0.9792}    & 0.9752$\ast$& \textbf{0.9789}   & 0.9752$\ast$& \textbf{0.9792} & 0.9755$\ast$& \textbf{0.9790} \\

8  & 0.5016$\ast$& \textbf{0.5260}  & 0.4996$\ast$& \textbf{0.5264}   & 0.4983$\ast$& \textbf{0.5256}    & 0.4953$\ast$& \textbf{0.5241}   & 0.4956$\ast$& \textbf{0.5202} & 0.4953$\ast$& \textbf{0.5209} \\

9  & 0.8574$\ast$& \textbf{0.8676}  & 0.8576$\ast$& \textbf{0.8680}   & 0.8577$\ast$& \textbf{0.8673}    & 0.8594$\ast$& \textbf{0.8687}   & 0.8592$\ast$& \textbf{0.8683} & 0.8594$\ast$& \textbf{0.8683} \\

10 & 0.5891$\ast$& \textbf{0.6148}  & 0.5882$\ast$& \textbf{0.6173}   & 0.5874$\ast$& \textbf{0.6171}    & 0.5971$\ast$& \textbf{0.6208}   & 0.5975$\ast$& \textbf{0.6204} & 0.5992$\ast$& \textbf{0.6196} \\

11 & 0.9111$\ast$& \textbf{0.9150}  & 0.9119$\ast$& \textbf{0.9158}   & 0.9121$\ast$& \textbf{0.9160}    & 0.9130$\ast$& \textbf{0.9155}   & 0.9134$\ast$& \textbf{0.9165} & 0.9138$\ast$& \textbf{0.9164} \\

12 & 0.9093$\ast$& \textbf{0.9309}  & 0.9102$\ast$& \textbf{0.9306}   & 0.9103$\ast$& \textbf{0.9302}    & 0.9119$\ast$& \textbf{0.9294}   & 0.9123$\ast$& \textbf{0.9298} & 0.9119$\ast$& \textbf{0.9303} \\

13 & 0.8789$\ast$  & \textbf{0.8801}       & 0.8791$\ast$  & \textbf{0.8801}         & 0.8790$\ast$  & \textbf{0.8801}  & 0.8792$\ast$  & \textbf{0.8800}       & 0.8792 $\ast$  & \textbf{0.8800}         & 0.8792$\ast$  & \textbf{0.8800}\\
\hline
\end{tabular*}
\vspace{-25pt}
\end{table*}

\begin{table*}[!ht]
\caption{Comparison of the $F_1$-score for SRS and RSS under exponential loss for different \(T\) values, with analysis of results via voting and mean methods under different settings}
\label{table_exp_f1}
\tiny
\setlength{\tabcolsep}{1pt} 
\begin{tabular*}{\linewidth}{@{\extracolsep{\fill}}l*{12}{c}}
\hline
\multicolumn{1}{l}{\multirow{2}{*}{Id}}& 
\multicolumn{6}{c}{exp\_vote\_Accuracy}& 
\multicolumn{6}{c}{exp\_mean\_Accuracy} \\
\cmidrule(lr){2-7} \cmidrule(lr){8-13}
& \multicolumn{2}{c}{$T=31$}
& \multicolumn{2}{c}{$T=41$}
& \multicolumn{2}{c}{$T=51$}
& \multicolumn{2}{c}{$T=31$}
& \multicolumn{2}{c}{$T=41$}
& \multicolumn{2}{c}{$T=51$} \\
\cmidrule(lr){2-3} \cmidrule(lr){4-5} \cmidrule(lr){6-7} 
\cmidrule(lr){8-9} \cmidrule(lr){10-11} \cmidrule(lr){12-13}
 & SRS & RSS  & SRS & RSS  & SRS & RSS & SRS & RSS  & SRS & RSS  & SRS & RSS \\
\hline
1  & 0.8032$\ast$& \textbf{0.8252} 	 & 0.8045$\ast$& \textbf{0.8257} 	  & 0.8038$\ast$& \textbf{0.8255}        & 0.9003$\ast$& \textbf{0.9085} 	  & 0.9014$\ast$& \textbf{0.9086} 	& 0.9010$\ast$& \textbf{0.9088}\\

2  & 0.8076$\ast$& \textbf{0.8397} 	 & 0.8083$\ast$& \textbf{0.8370} 	  & 0.8097$\ast$& \textbf{0.8383}        & 0.8544$\ast$& \textbf{0.8792} 	  & 0.8537$\ast$& \textbf{0.8781} 	& 0.8546$\ast$& \textbf{0.8776}\\

3  & 0.6408$\ast$& \textbf{0.6566} 	 & 0.6411$\ast$& \textbf{0.6572} 	  & 0.6414$\ast$& \textbf{0.6578}        & 0.6518$\ast$& \textbf{0.6623} 	  & 0.6515$\ast$& \textbf{0.6625} 	& 0.6514$\ast$& \textbf{0.6634}\\

4  & 0.5570$\ast$& \textbf{0.6048} 	 & 0.5533$\ast$& \textbf{0.6026} 	  & 0.5508$\ast$& \textbf{0.6009}        & 0.7606$\ast$& \textbf{0.7724} 	  & 0.7606$\ast$& \textbf{0.7722} 	& 0.7622$\ast$& \textbf{0.7717}\\

5  & 0.8441$\ast$& \textbf{0.8487} 	 & 0.8444$\ast$& \textbf{0.8492} 	  & 0.8444$\ast$& \textbf{0.8489}        & 0.8647$\ast$& \textbf{0.8685} 	  & 0.8649$\ast$& \textbf{0.8685} 	& 0.8650$\ast$& \textbf{0.8686}\\

6  & 0.9719$\ast$& \textbf{0.9758} 	 & 0.9720$\ast$& \textbf{0.9757} 	  & 0.9721$\ast$& \textbf{0.9756}        & 0.9713$\ast$& \textbf{0.9753} 	  & 0.9714$\ast$& \textbf{0.9755} 	& 0.9716$\ast$& \textbf{0.9755}\\

7  & 0.9741$\ast$& \textbf{0.9793} 	 & 0.9738$\ast$& \textbf{0.9792} 	  & 0.9743$\ast$& \textbf{0.9788}        & 0.9753$\ast$& \textbf{0.9784} 	  & 0.9754$\ast$& \textbf{0.9788} 	& 0.9756$\ast$& \textbf{0.9787}\\

8  & 0.4746$\ast$& \textbf{0.4990} 	 & 0.4711$\ast$& \textbf{0.4970}      & 0.4719$\ast$& \textbf{0.4979}        & 0.5012$\ast$& \textbf{0.5204} 	  & 0.5002$\ast$& \textbf{0.5181}   & 0.4992$\ast$& \textbf{0.5193}\\

9  & 0.8567$\ast$& \textbf{0.8673} 	 & 0.8576$\ast$& \textbf{0.8680} 	  & 0.8577$\ast$& \textbf{0.8673}        & 0.8596$\ast$& \textbf{0.8686} 	  & 0.8592$\ast$& \textbf{0.8683} 	& 0.8594$\ast$& \textbf{0.8683}\\

10 & 0.5157$\ast$& \textbf{0.5739} 	 & 0.5103$\ast$&\textbf{0.5773} 	  & 0.5099$\ast$& \textbf{0.5768}        & 0.6017$\ast$& \textbf{0.6174} 	  & 0.6014$\ast$& \textbf{0.6169} 	& 0.6033$\ast$& \textbf{0.6163}\\

11 & 0.9109$\ast$& \textbf{0.9145} 	 & 0.9115$\ast$& \textbf{0.9154} 	  & 0.9118$\ast$& \textbf{0.9156}        & 0.9129$\ast$& \textbf{0.9156} 	  & 0.9132$\ast$& \textbf{0.9165} 	& 0.9136$\ast$& \textbf{0.9164}\\

12 & 0.8822$\ast$& \textbf{0.9110} 	 & 0.8834$\ast$& \textbf{0.9110} 	  & 0.8846$\ast$& \textbf{0.9093}        & 0.9121$\ast$& \textbf{0.9279} 	  & 0.9124$\ast$& \textbf{0.9283} 	& 0.9117$\ast$& \textbf{0.9288}\\

13 & 0.8789$\ast$  & \textbf{0.8801}       & 0.8792$\ast$  & \textbf{0.8799}         & 0.8789$\ast$  & \textbf{0.8801}  & 0.8792$\ast$  & \textbf{0.8799}       & 0.8792$\ast$  & \textbf{0.8798}         & 0.8792$\ast$  & \textbf{0.8798}\\
\hline
\end{tabular*}
\vspace{-25pt}
\end{table*}

\begin{table*}[!ht]
\caption{Comparison of SRS and RSS accuracy under logistic loss for different \(T\) values, with analysis of results via voting and mean methods under different settings}
\label{table_log_acc}
\tiny
\setlength{\tabcolsep}{1pt} 
\begin{tabular*}{\linewidth}{@{\extracolsep{\fill}}l*{12}{c}}
\hline
\multicolumn{1}{l}{\multirow{2}{*}{
\begin{tabular}[c]{@{}l@{}}Id\end{tabular}}}
& \multicolumn{6}{c}{log\_vote\_Accuracy}
& \multicolumn{6}{c}{log\_mean\_Accuracy} \\
\cmidrule(l){2-7}\cmidrule(l){8-13}
&\multicolumn{2}{c}{$T$ = 31}
&\multicolumn{2}{c}{$T$ = 41}
&\multicolumn{2}{c}{$T$ = 51}
& \multicolumn{2}{c}{$T$ = 31}
&\multicolumn{2}{c}{$T$ = 41}
&\multicolumn{2}{c}{$T$ = 51} \\
\cmidrule(l){2-3}\cmidrule(l){4-5}\cmidrule(l){6-7}\cmidrule(l){8-9}\cmidrule(l){10-11}\cmidrule(l){12-13}
\multicolumn{1}{l}{}
  & SRS & RSS  & SRS & RSS  & SRS & RSS & SRS & RSS  & SRS & RSS  & SRS & RSS \\
\hline
1  & 0.9007$\ast$& \textbf{0.9106}   & 0.9001$\ast$& \textbf{0.9114} & 0.9005$\ast$& \textbf{0.9112}           & 0.9031$\ast$& \textbf{0.9132} & 0.9041$\ast$& \textbf{0.9123}    & 0.9043$\ast$& \textbf{0.9124}\\

2  & 0.8555$\ast$& \textbf{0.8806}   & 0.8564$\ast$& \textbf{0.8782} & 0.8553$\ast$& \textbf{0.8777}           & 0.8606$\ast$& \textbf{0.8884} & 0.8619$\ast$& \textbf{0.8890}    & 0.8619$\ast$& \textbf{0.8877}\\

3  & 0.6481$\ast$& \textbf{0.6660}   & 0.6484$\ast$& \textbf{0.6657} & 0.6494$\ast$& \textbf{0.6657}           & 0.6538$\ast$& \textbf{0.6646} & 0.6536$\ast$& \textbf{0.6659}    & 0.6544$\ast$& \textbf{0.6653}\\

4  & 0.7570$\ast$& \textbf{0.7853}   & 0.7578$\ast$& \textbf{0.7857} & 0.7586$\ast$& \textbf{0.7863}           & 0.7581$\ast$& \textbf{0.7865} & 0.7586$\ast$& \textbf{0.7876}    & 0.7588$\ast$& \textbf{0.7863}\\

5  & 0.8672$\ast$& \textbf{0.8719}   & 0.8478$\ast$& \textbf{0.8526} & 0.8674$\ast$& \textbf{0.8722}           & 0.8667$\ast$& \textbf{0.8710} & 0.8668$\ast$& \textbf{0.8704}    & 0.8665$\ast$& \textbf{0.8710}\\

6  & 0.9703$\ast$& \textbf{0.9740}   & 0.9702$\ast$& \textbf{0.9741} & 0.9702$\ast$& \textbf{0.9741}           & 0.9710$\ast$& \textbf{0.9747} & 0.9710$\ast$& \textbf{0.9748}    & 0.9712$\ast$& \textbf{0.9747}\\

7  & 0.9770$\ast$& \textbf{0.9833}   & 0.9773$\ast$& \textbf{0.9832} & 0.9772$\ast$& \textbf{0.9833}           & 0.9758$\ast$& \textbf{0.9821} & 0.9757$\ast$& \textbf{0.9821}    & 0.9758$\ast$& \textbf{0.9819}\\

8  & 0.4981$\ast$& \textbf{0.5186}   & 0.4964$\ast$& \textbf{0.5194} & 0.4969$\ast$& \textbf{0.5201}           & 0.4941$\ast$& \textbf{0.5142} & 0.4931$\ast$& \textbf{0.5142} 	  & 0.4946$\ast$& \textbf{0.5138}\\

9  & 0.8601$\ast$& \textbf{0.8707}   & 0.8604$\ast$& \textbf{0.8706} & 0.8601$\ast$& \textbf{0.8707}           & 0.8602$\ast$& \textbf{0.8689} & 0.8602$\ast$& \textbf{0.8686}    & 0.8602$\ast$& \textbf{0.8687}\\

10 & 0.5990$\ast$& \textbf{0.6230}   & 0.6006$\ast$& \textbf{0.6231} & 0.6005$\ast$& \textbf{0.6228}           & 0.6002$\ast$& \textbf{0.6212} & 0.5998$\ast$& \textbf{0.6204} 	  & 0.5995$\ast$& \textbf{0.6209}\\

11 & 0.9302$\ast$& \textbf{0.9347}   & 0.9308$\ast$& \textbf{0.9349} & 0.9308$\ast$& \textbf{0.9352}           & 0.9292$\ast$& \textbf{0.9340} & 0.9294$\ast$& \textbf{0.9342}    & 0.9297$\ast$& \textbf{0.9344}\\

12 & 0.9068$\ast$& \textbf{0.9258}   & 0.9065$\ast$& \textbf{0.9260} & 0.9062$\ast$& \textbf{0.9265}           & 0.9111$\ast$& \textbf{0.9302} & 0.9111$\ast$& \textbf{0.9297}    & 0.9107$\ast$& \textbf{0.9303}\\

13 & 0.8833$\ast$  & \textbf{0.8843}       & 0.8833$\ast$  & \textbf{0.8842}         & 0.8837$\ast$  & \textbf{0.8845}  & 0.8832$\ast$  & \textbf{0.8840}       & 0.8832$\ast$  & \textbf{0.8836}         & 0.8834$\ast$  & \textbf{0.8836}\\
\hline
\end{tabular*}
\vspace{-15pt}
\end{table*}

\begin{table*}[!ht]
\caption{Comparison of SRS and RSS $F_1$-score under logistic loss for different \(T\) values, with analysis of results via voting and mean methods under different settings}
\label{table_log_f1}
\tiny
\setlength{\tabcolsep}{1pt} 
\begin{tabular*}{\linewidth}{@{\extracolsep{\fill}}l*{12}{c}}
\hline
\multicolumn{1}{l}{\multirow{2}{*}{
\begin{tabular}[c]{@{}l@{}}Id\end{tabular}}}
& \multicolumn{6}{c}{log\_vote\_$F_{1}$-score}
& \multicolumn{6}{c}{log\_mean\_$F_{1}$-score} \\
\cmidrule(l){2-7}\cmidrule(l){8-13}
&\multicolumn{2}{c}{$T$ = 31}
&\multicolumn{2}{c}{$T$ = 41}
&\multicolumn{2}{c}{$T$ = 51}
& \multicolumn{2}{c}{$T$ = 31}
&\multicolumn{2}{c}{$T$ = 41}
&\multicolumn{2}{c}{$T$ = 51} \\
\cmidrule(l){2-3}\cmidrule(l){4-5}\cmidrule(l){6-7}\cmidrule(l){8-9}\cmidrule(l){10-11}\cmidrule(l){12-13}
\multicolumn{1}{l}{}
  & SRS & RSS  & SRS & RSS  & SRS & RSS & SRS & RSS  & SRS & RSS  & SRS & RSS \\
\hline
1  & 0.8119$\ast$& \textbf{0.8289} & 0.8107$\ast$& \textbf{0.8307}& 0.8115$\ast$& \textbf{0.8303}               & 0.9045$\ast$& \textbf{0.9110}& 0.9056$\ast$& \textbf{0.9111} & 0.9054$\ast$& \textbf{0.9113}  \\

2  & 0.8180$\ast$& \textbf{0.8450} & 0.8190$\ast$& \textbf{0.8425}& 0.8180$\ast$& \textbf{0.8408}               & 0.8634$\ast$& \textbf{0.8887}& 0.8645$\ast$& \textbf{0.8896} & 0.8645$\ast$& \textbf{0.8877}  \\

3  & 0.6394$\ast$& \textbf{0.6572} & 0.6399$\ast$& \textbf{0.6573}& 0.6410$\ast$& \textbf{0.6575}               & 0.6544$\ast$& \textbf{0.6638}& 0.6546$\ast$& \textbf{0.6650} & 0.6552$\ast$& \textbf{0.6645}  \\

4  & 0.5536$\ast$& \textbf{0.6044} & 0.5536$\ast$& \textbf{0.6054}& 0.5563$\ast$& \textbf{0.6051}               & 0.7596$\ast$& \textbf{0.7826}& 0.7606$\ast$& \textbf{0.7834} & 0.7608$\ast$& \textbf{0.7830}  \\

5  & 0.8478$\ast$& \textbf{0.8526} & 0.8478$\ast$& \textbf{0.8527}& 0.8477$\ast$& \textbf{0.8527}               & 0.8668$\ast$& \textbf{0.8704}& 0.8667$\ast$& \textbf{0.8705} & 0.8666$\ast$& \textbf{0.8705}  \\

6  & 0.9702$\ast$& \textbf{0.9738} & 0.9701$\ast$& \textbf{0.9739}& 0.9701$\ast$& \textbf{0.9739}               & 0.9711$\ast$& \textbf{0.9747}& 0.9712$\ast$& \textbf{0.9748} & 0.9713$\ast$& \textbf{0.9747}  \\

7  & 0.9777$\ast$& \textbf{0.9834} & 0.9779$\ast$& \textbf{0.9834}& 0.9777$\ast$& \textbf{0.9835}               & 0.9758$\ast$& \textbf{0.9821}& 0.9757$\ast$& \textbf{0.9821} & 0.9758$\ast$& \textbf{0.9819} \\

8  & 0.4728$\ast$& \textbf{0.4925} & 0.4715$\ast$& \textbf{0.4933}& 0.4714$\ast$& \textbf{0.4933}               & 0.4944$\ast$& \textbf{0.5141}& 0.4932$\ast$& \textbf{0.5142} & 0.4949$\ast$& \textbf{0.5141}  \\

9  & 0.8597$\ast$& \textbf{0.8703} & 0.8601$\ast$& \textbf{0.8702}& 0.8597$\ast$& \textbf{0.8704}               & 0.8610$\ast$& \textbf{0.8685}& 0.8610$\ast$& \textbf{0.8686} & 0.8609$\ast$& \textbf{0.8687}  \\

10 & 0.5300$\ast$& \textbf{0.5771} & 0.5255$\ast$& \textbf{0.5774}& 0.5328$\ast$& \textbf{0.5775}               & 0.6035$\ast$& \textbf{0.6173}& 0.6025$\ast$& \textbf{0.6164} & 0.6028$\ast$& \textbf{0.6162}  \\

11 & 0.9300$\ast$& \textbf{0.9343} & 0.9308$\ast$& \textbf{0.9344}& 0.9308$\ast$& \textbf{0.9347}               & 0.9294$\ast$& \textbf{0.9341}& 0.9297$\ast$& \textbf{0.9343} & 0.9299$\ast$& \textbf{0.9345} \\

12 & 0.8851$\ast$& \textbf{0.9104} & 0.8829$\ast$& \textbf{0.9101}& 0.8827$\ast$& \textbf{0.9117}               & 0.9121$\ast$& \textbf{0.9296}& 0.9123$\ast$& \textbf{0.9287} & 0.9121$\ast$& \textbf{0.9294}  \\

13 & 0.8833$\ast$  & \textbf{0.8842}       & 0.8833$\ast$   & \textbf{0.8842}         & 0.8837  & \textbf{0.8845} & 0.8832$\ast$  & \textbf{0.8839}       & 0.8833   & \textbf{0.8837}         & 0.8834   & \textbf{0.8836}  \\
\hline
\end{tabular*}
\vspace{-10pt}
\end{table*}

\begin{figure}[!h]
\centering
\subfloat[ExpL Accuracy]{\includegraphics[width=0.23\textwidth]{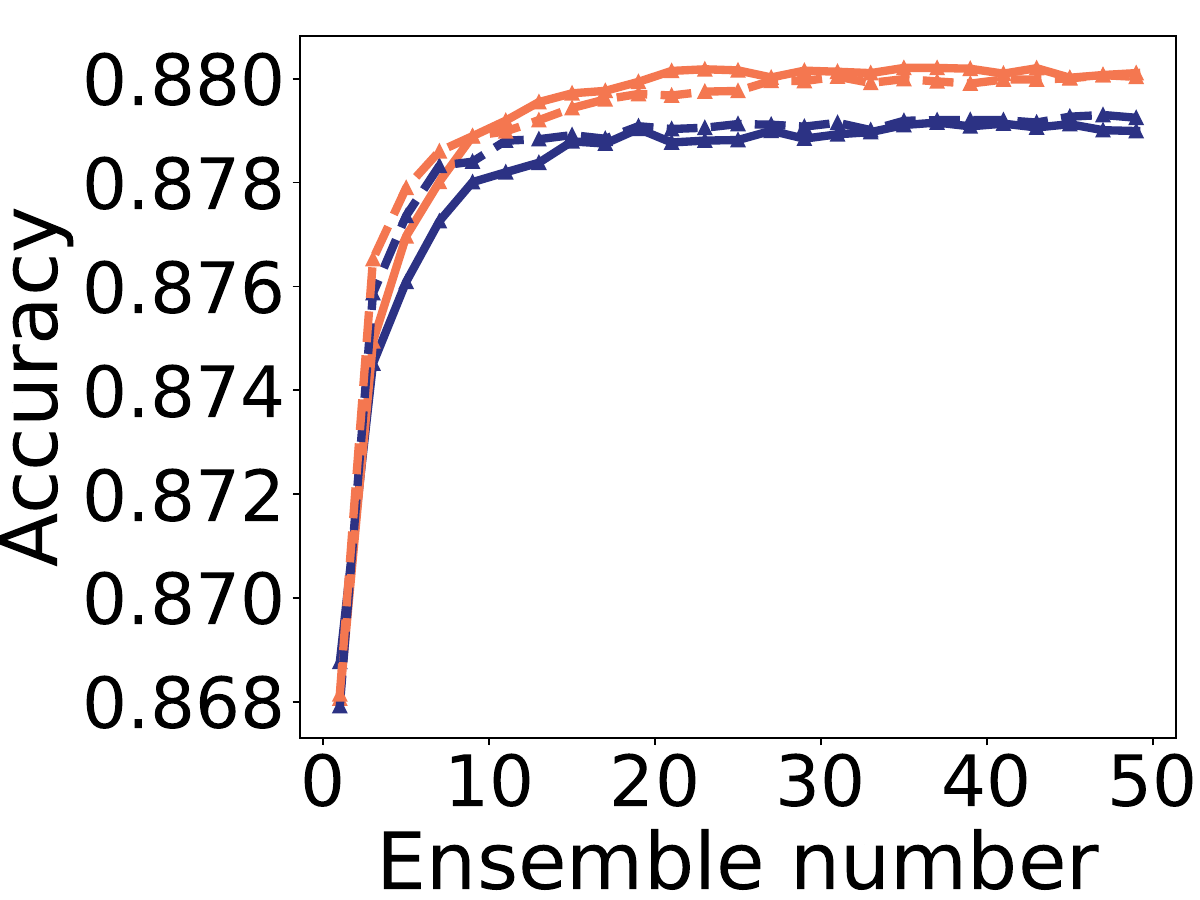}}
\subfloat[ExpL $F_{1}$-score]{\includegraphics[width=0.23\textwidth]{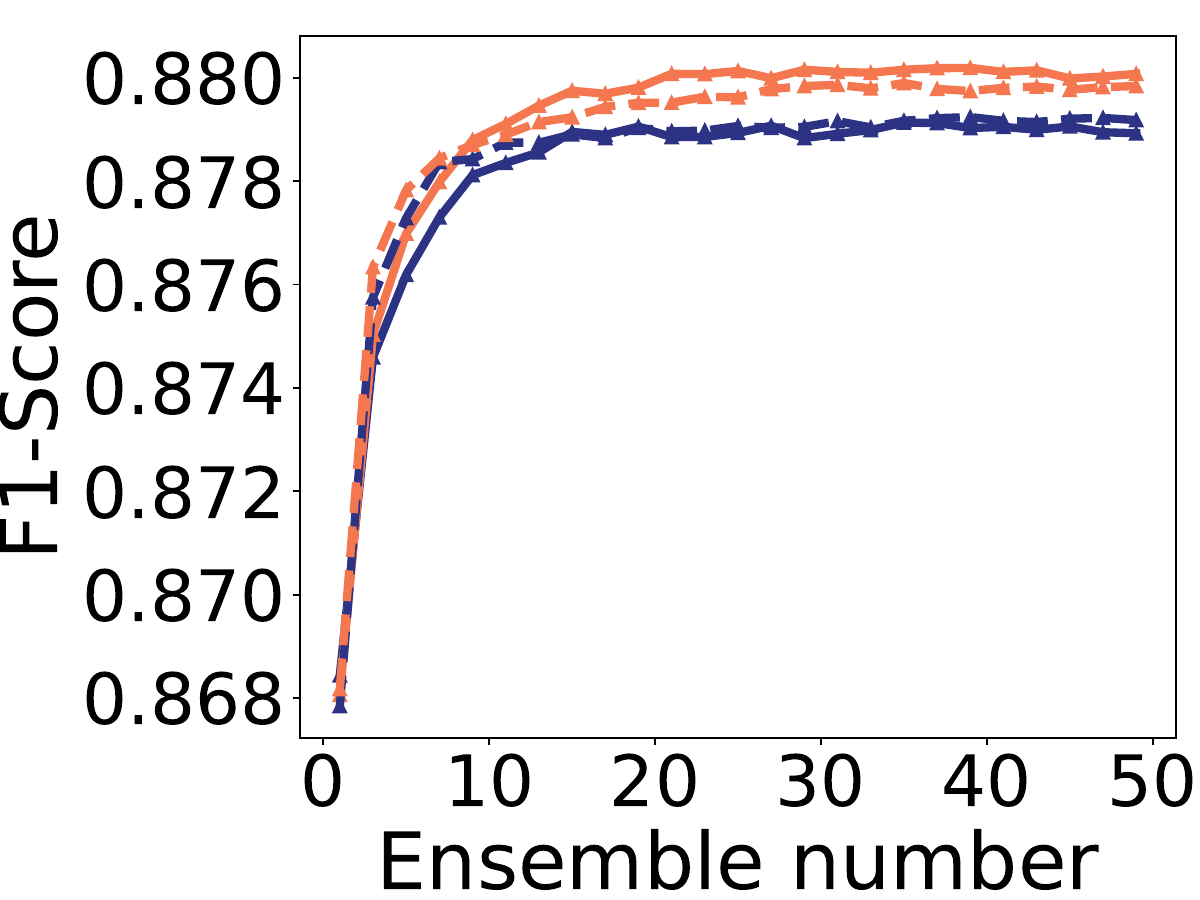}}  
\subfloat[LogL Accuracy]{\includegraphics[width=0.23\textwidth]{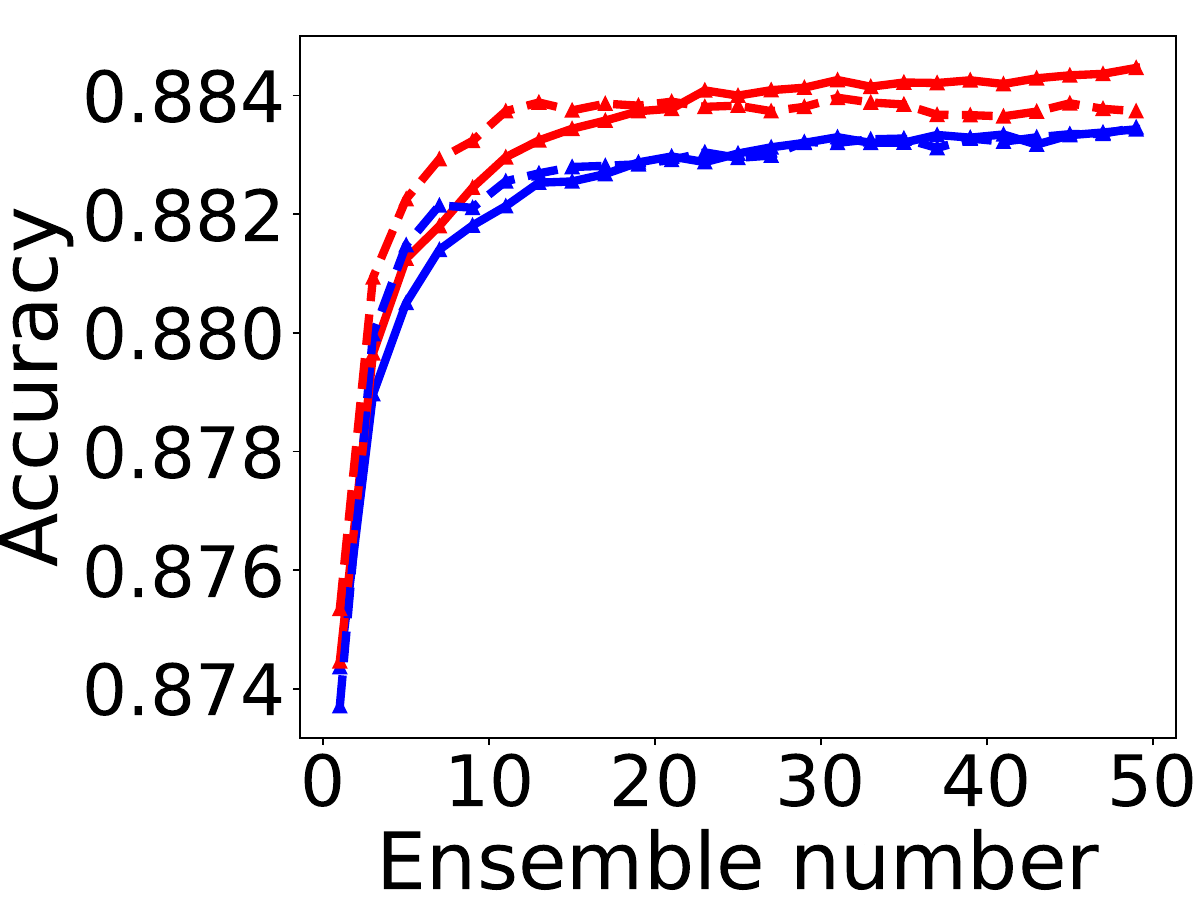}}
\subfloat[LogL $F_{1}$-score]{\includegraphics[width=0.23\textwidth]{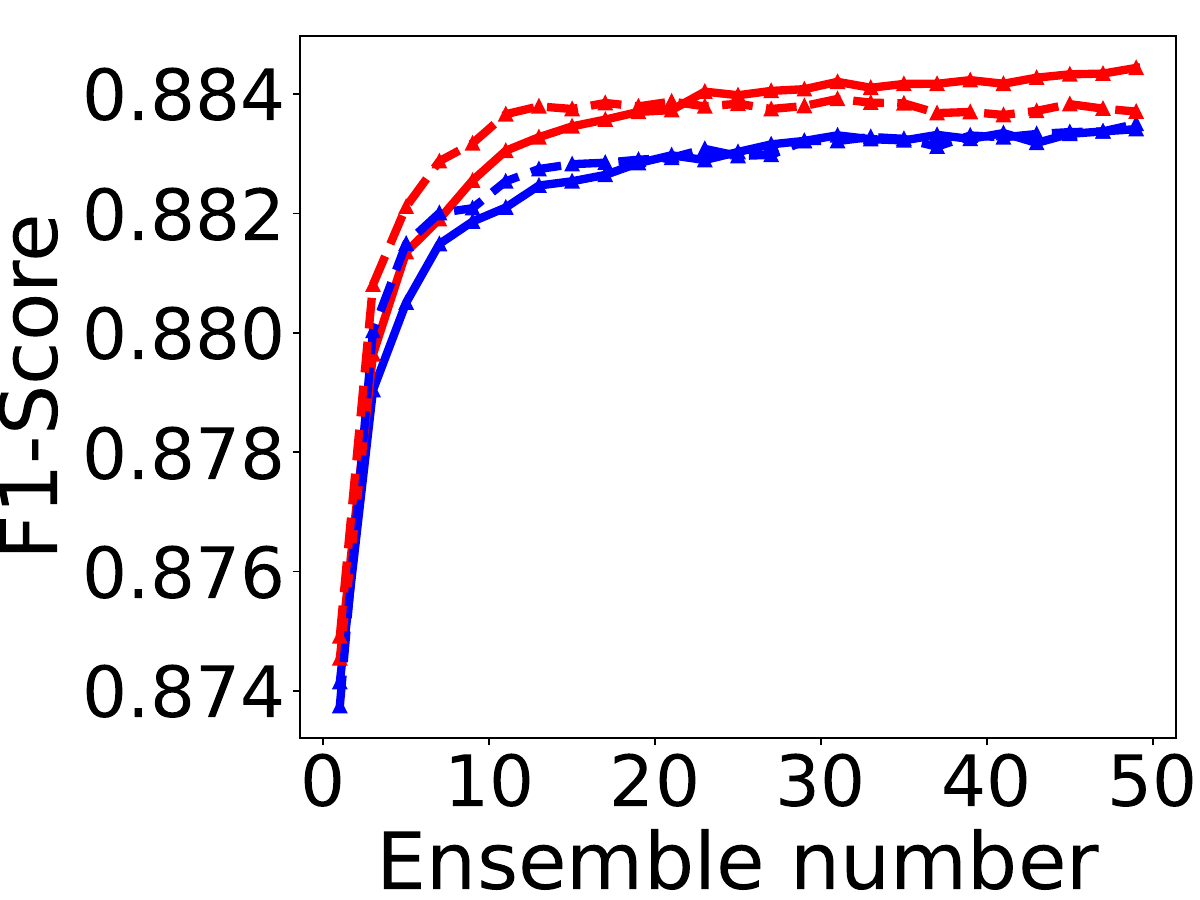}}\\
\subfloat{\includegraphics[width=0.47\textwidth]{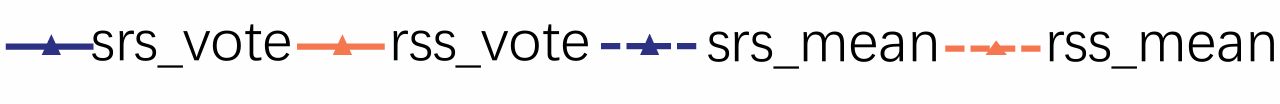}}
\subfloat{\includegraphics[width=0.47\textwidth]{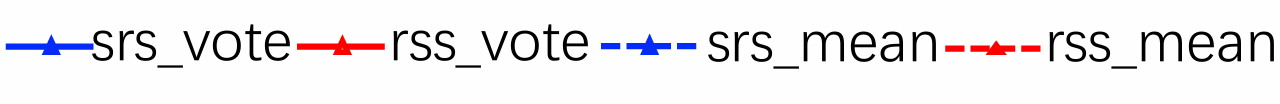}}
\caption{The Experimental Results of the Effect of the tree Number on the CIFAR-10 dataset.}
\label{image-plot}
\vspace{-15pt}
\end{figure}

\begin{figure*}[!h]
\centering
\subfloat[Data 1]{\includegraphics[width=0.2\textwidth]{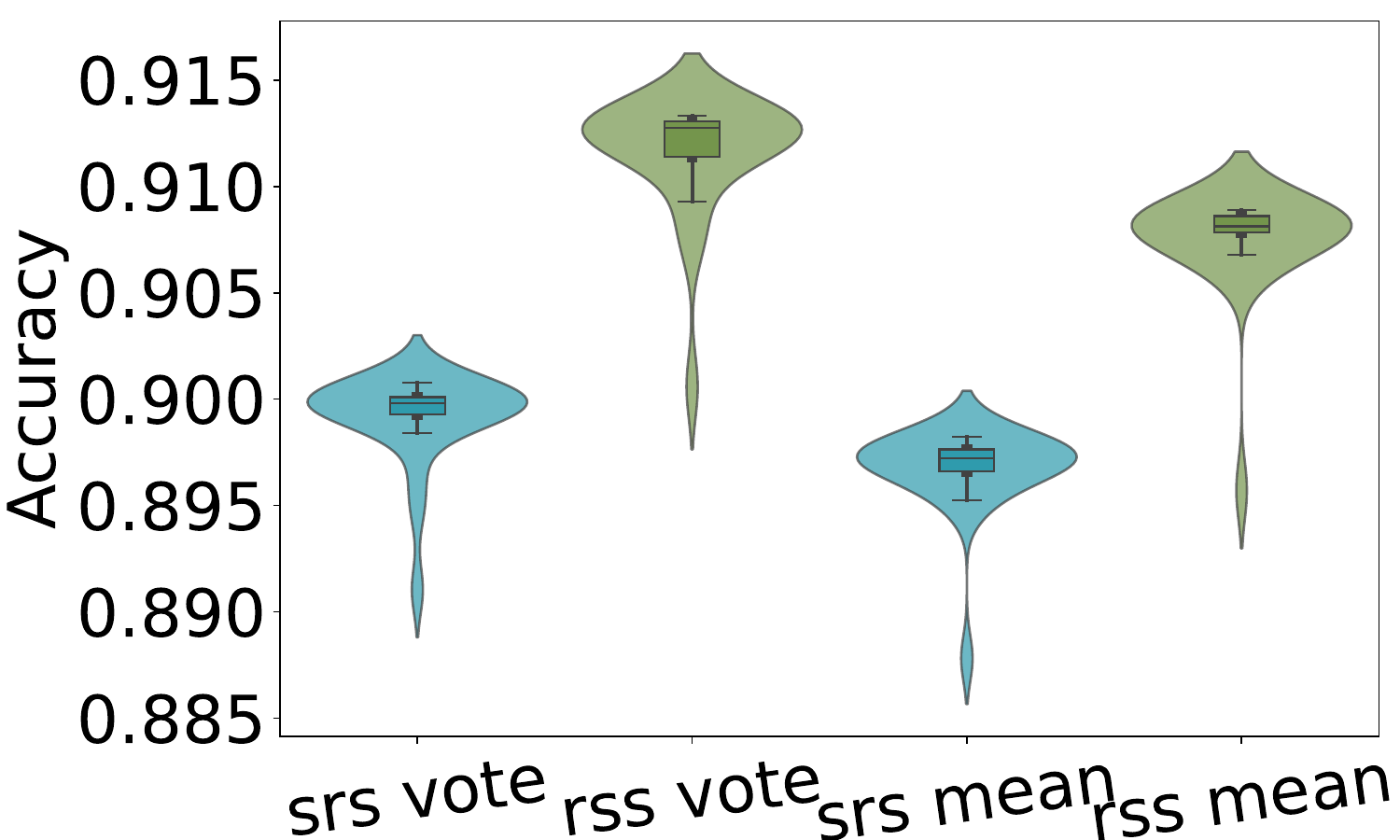}}
\subfloat[Data 2]{\includegraphics[width=0.2\textwidth]{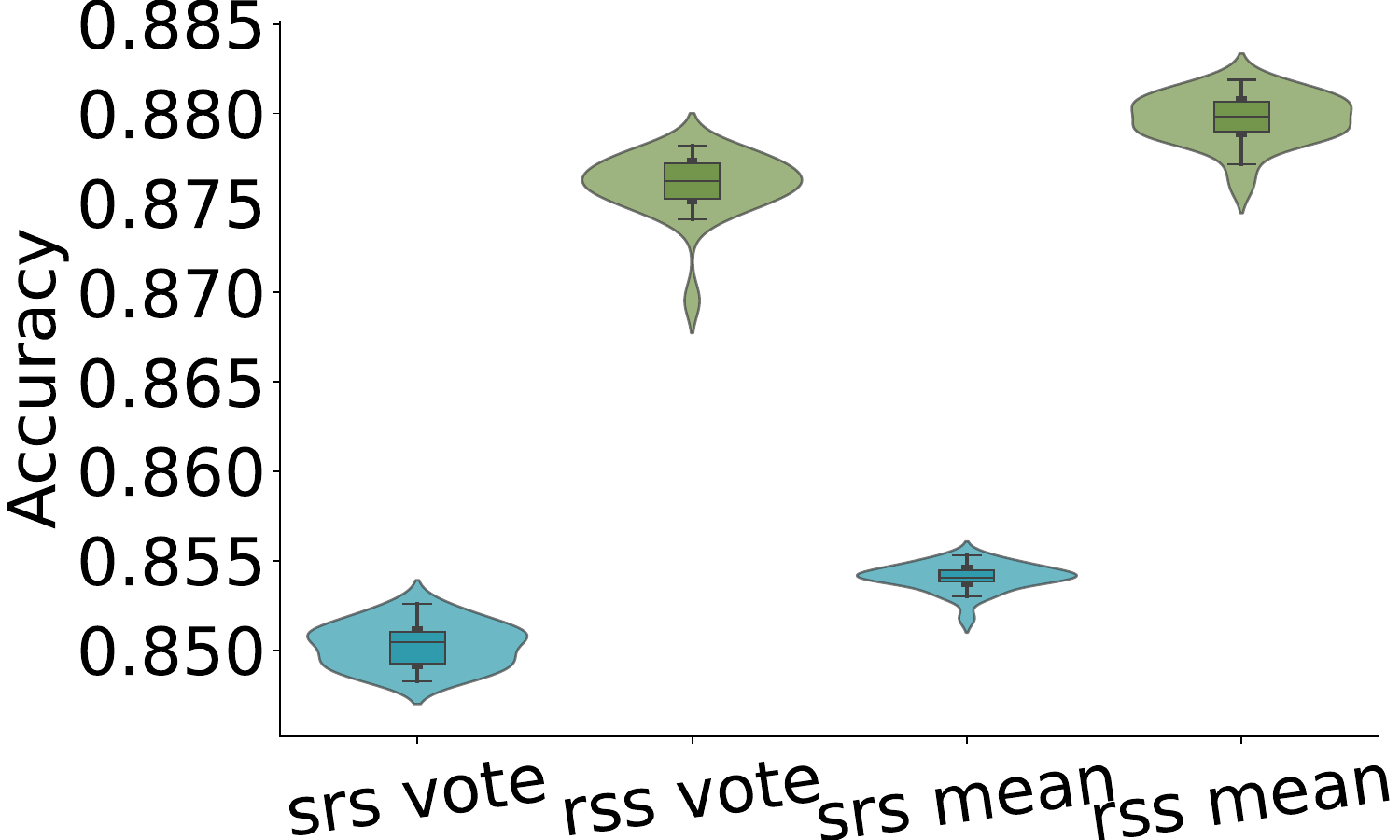}}    
\subfloat[Data 3]{\includegraphics[width=0.2\textwidth]{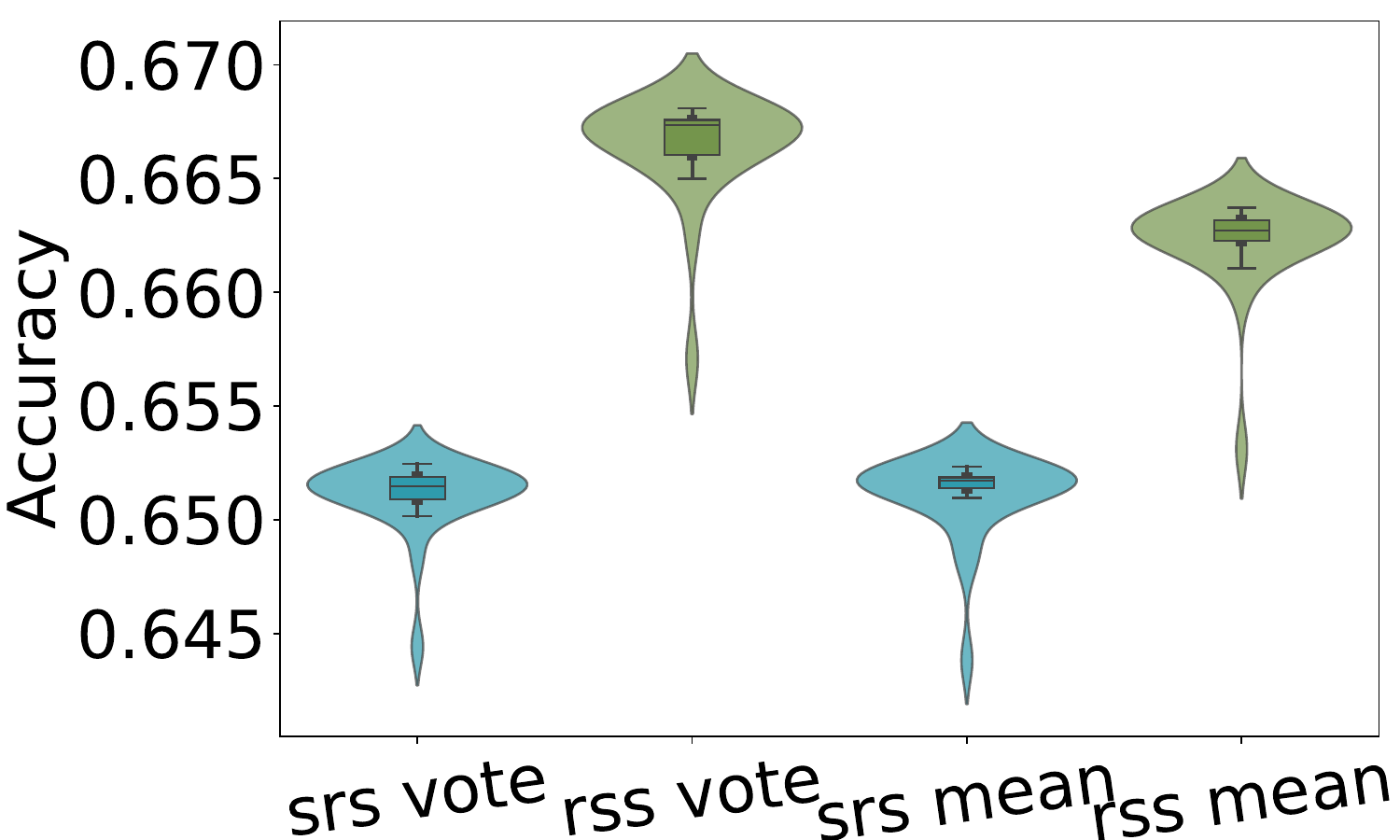}}
\subfloat[Data 4]{\includegraphics[width=0.2\textwidth]{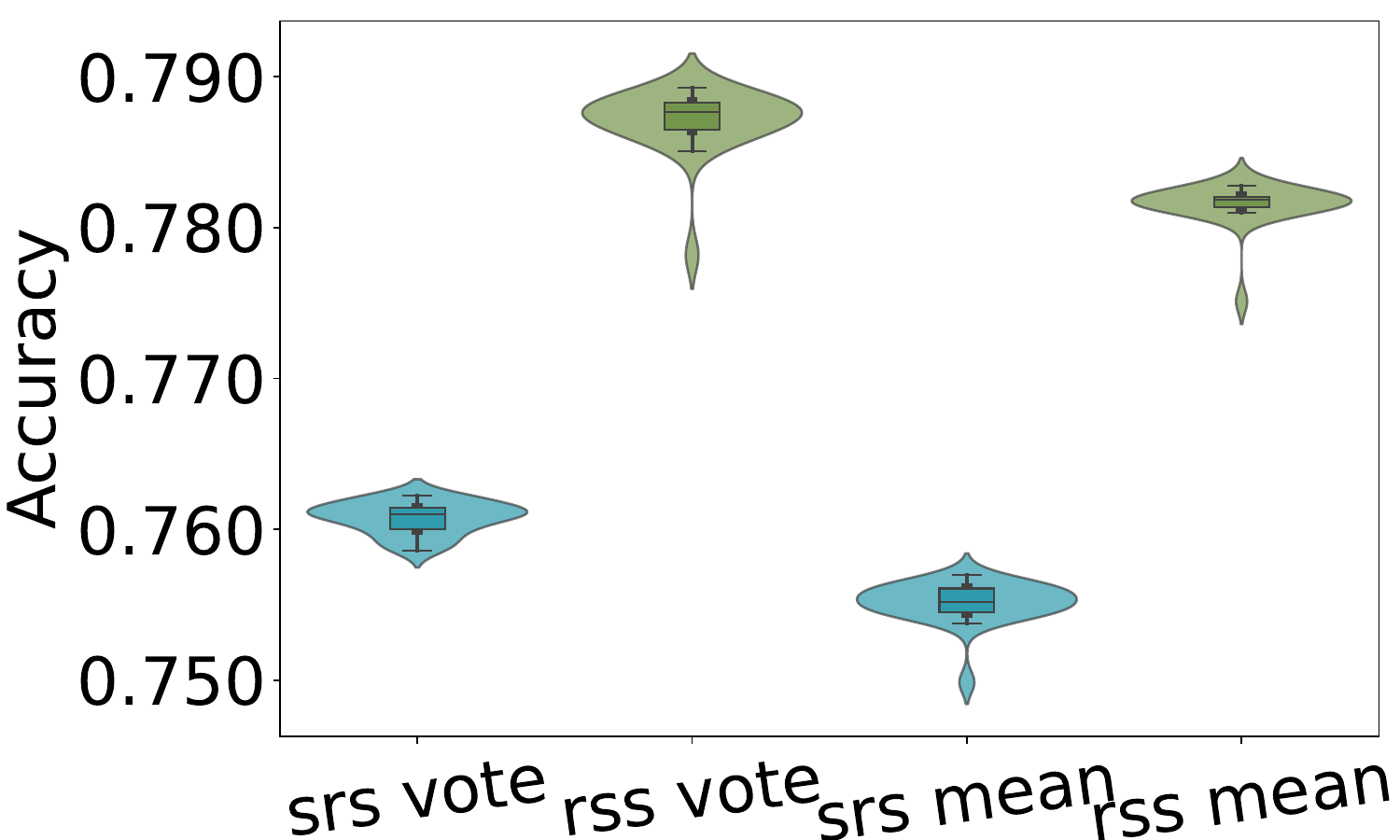}}
\subfloat[Data 5]{\includegraphics[width=0.2\textwidth]{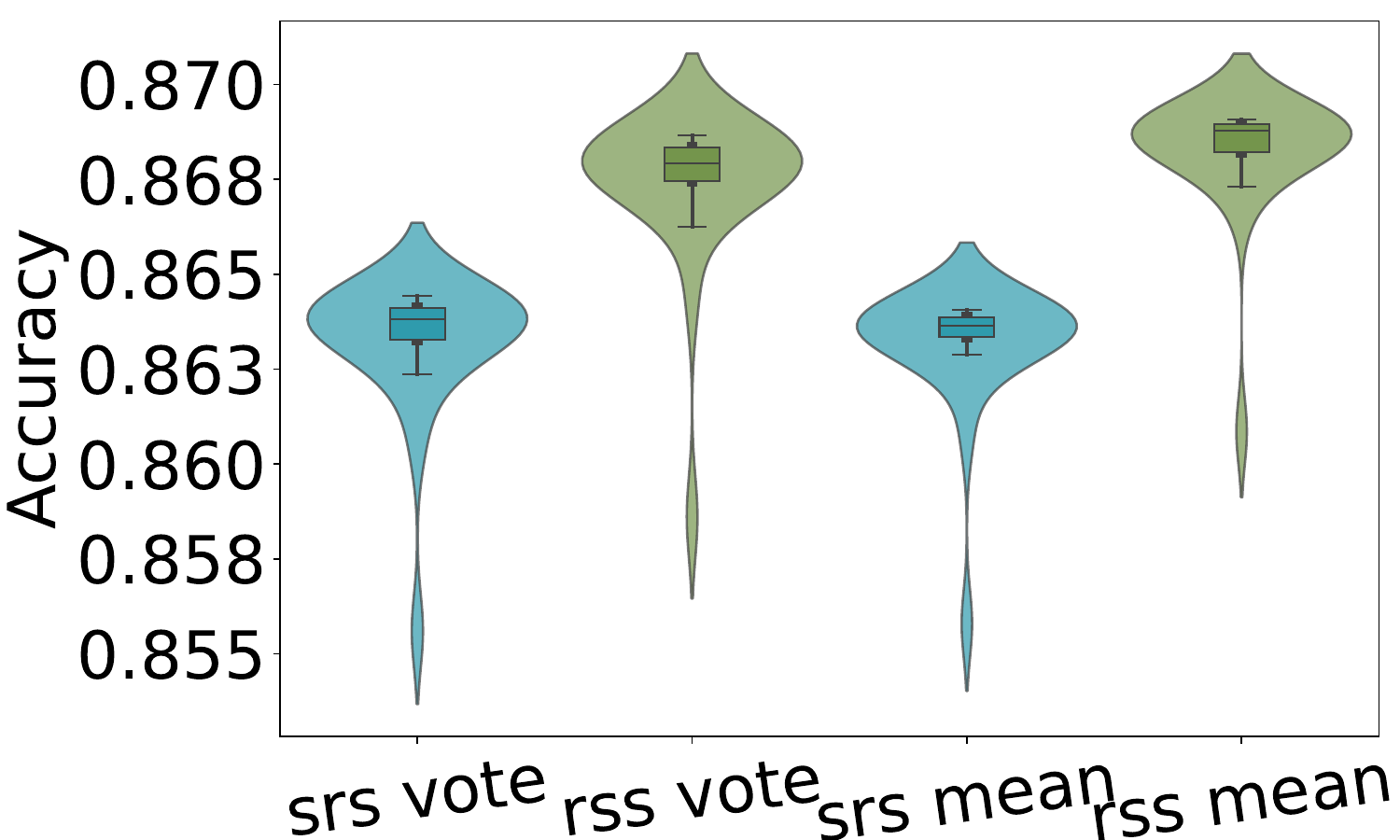}}
\caption{The Violin Diagram on the top five datasets.}
\label{box+image}
\vspace{-15pt}
\end{figure*}

  
\textbf{Ensemble Number Analysis.} From Table \ref{table_exp_acc}-Table \ref{table_log_f1}, it can be seen that RSS-MLP outperforms SRS-MLP under different number of base learners. We also show the performance curves of RSS-MLP and SRS-MLP when the number of ensembles increases on the CIFAR-10 dataset. The experimental results are shown in Figure \ref{image-plot}. As shown by Figure \ref{image-plot}, the RSS-MLP curve consistently outperform that of SRS-MLP across all classifier numbers. This finding underscores the efficacy of RSS-MLP with respect to both $F_1$-score and accuracy, particularly as the number of base classifiers increases. The solid line in the figures represents a voting-based integration strategy, while the dashed line indicates an average-based integration strategy. It is evident from these figures that, in most cases, the integration strategy based on mean strategy outperforms the voting strategy. In addition, the integrated approach consistently outperforms a single MLP approach (the position where the ensemble number is 1).

\begin{wrapfigure}[10]{r}{0.3\textwidth}
\vspace{-1.5em}
 \begin{minipage}{0.3\textwidth}
        \centering
        \includegraphics[width=0.98\linewidth]{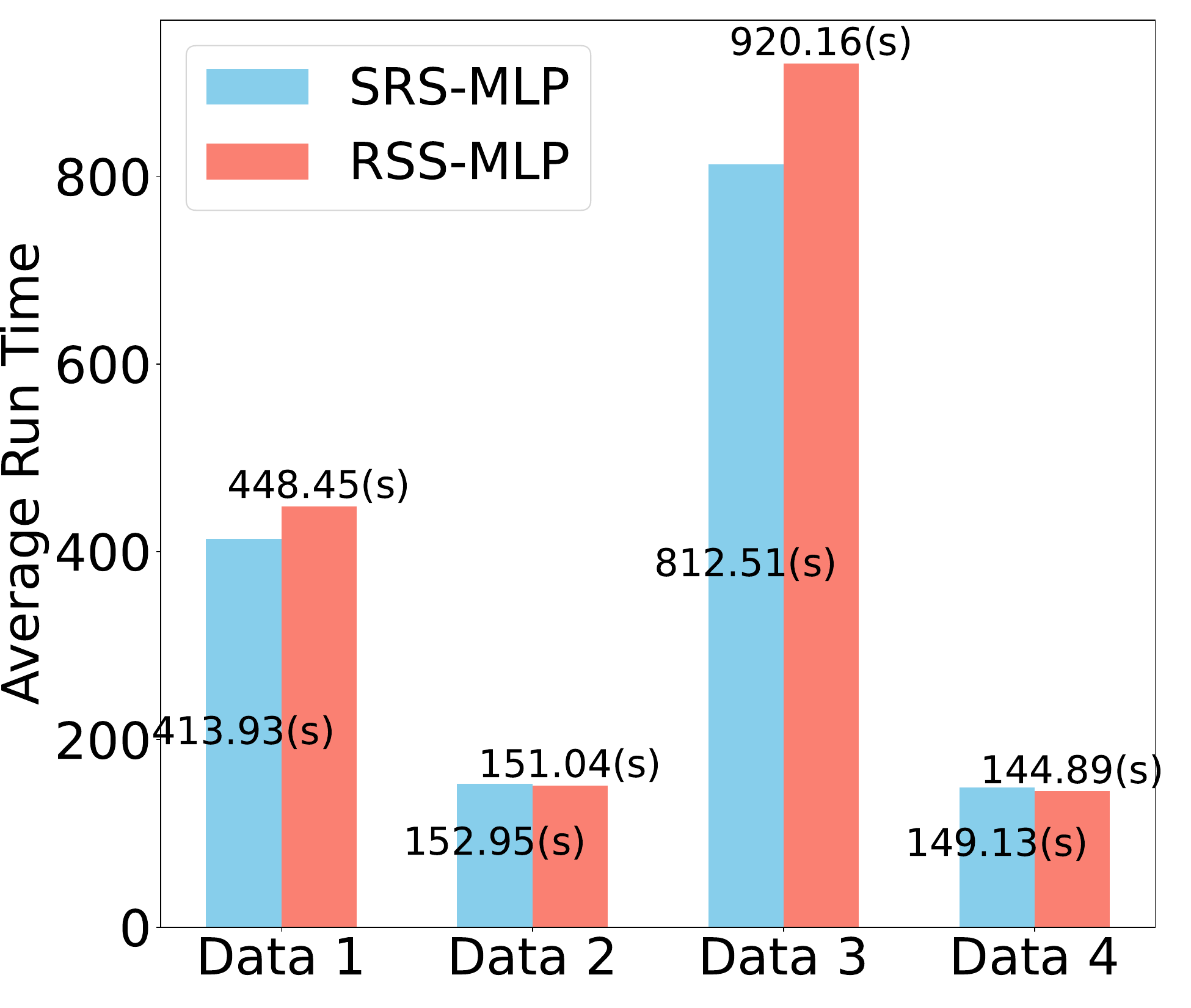}
        \caption{Time Consumption}
        \label{time_analysis}
    \end{minipage}
\end{wrapfigure} 

\textbf{Violin Diagram.} To show the distribution of the performance of multiple experiments, we show the violin diagram of the accuracy of the results of multiple experiments for the first five data under the Expl loss. The results are shown in Figure \ref{box+image}. From Figure \ref{box+image}, it can be seen that the stability of RSS-MLP and SRS-MLP is comparable, while RSS-MLP consistently outperforms SRS-MLP on all the five datasets.


\textbf{Time and Memory Consumption Analysis.} As for memory consumption, RSS-MLP only needs an extra $k\times k$ cache for order structure. As for time consumption, we show the time consumption on the first five data. As shown in Figure \ref{time_analysis}, it can be seen that RSS-MLP does not generate significant additional time overhead.

\begin{figure*}[!h]
\centering
\subfloat[ExpL Vote K]{\includegraphics[width=0.25\textwidth]{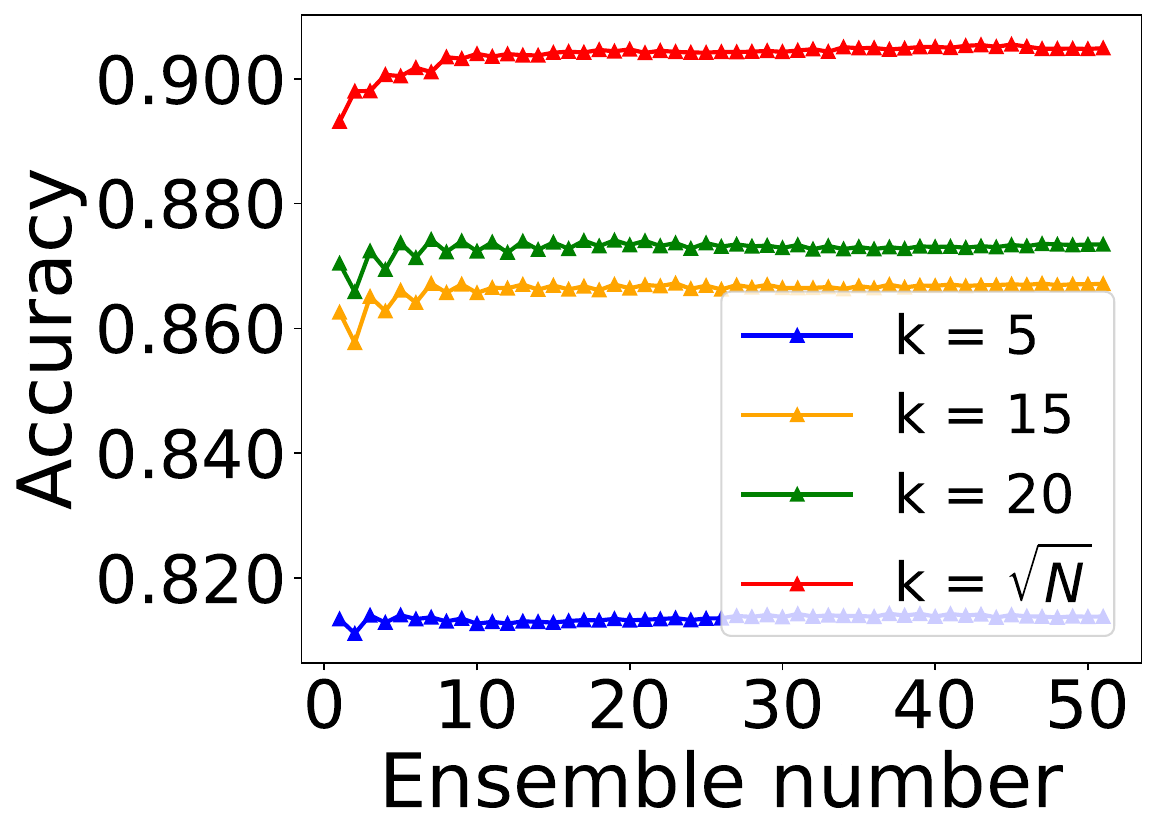}}
\subfloat[ExpL Mean K]{\includegraphics[width=0.25\textwidth]{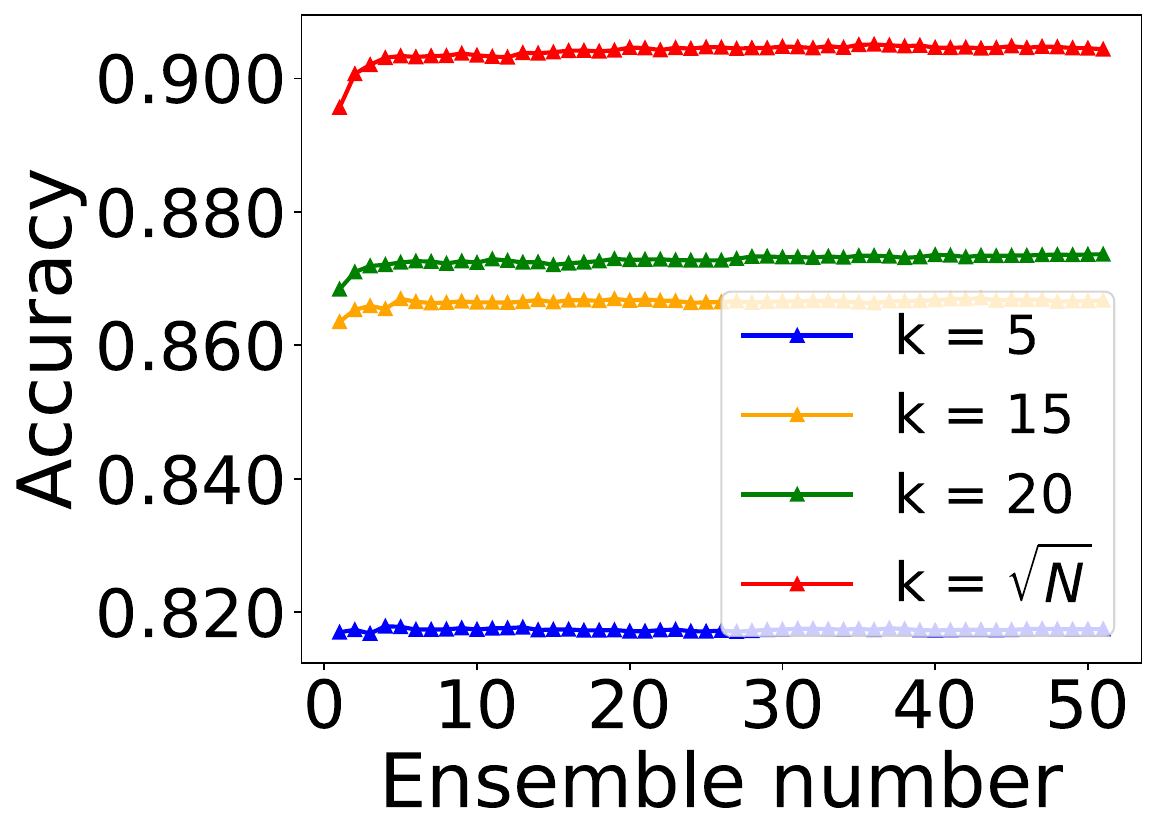}}   
\subfloat[LogL Vote K]{\includegraphics[width=0.25\textwidth]{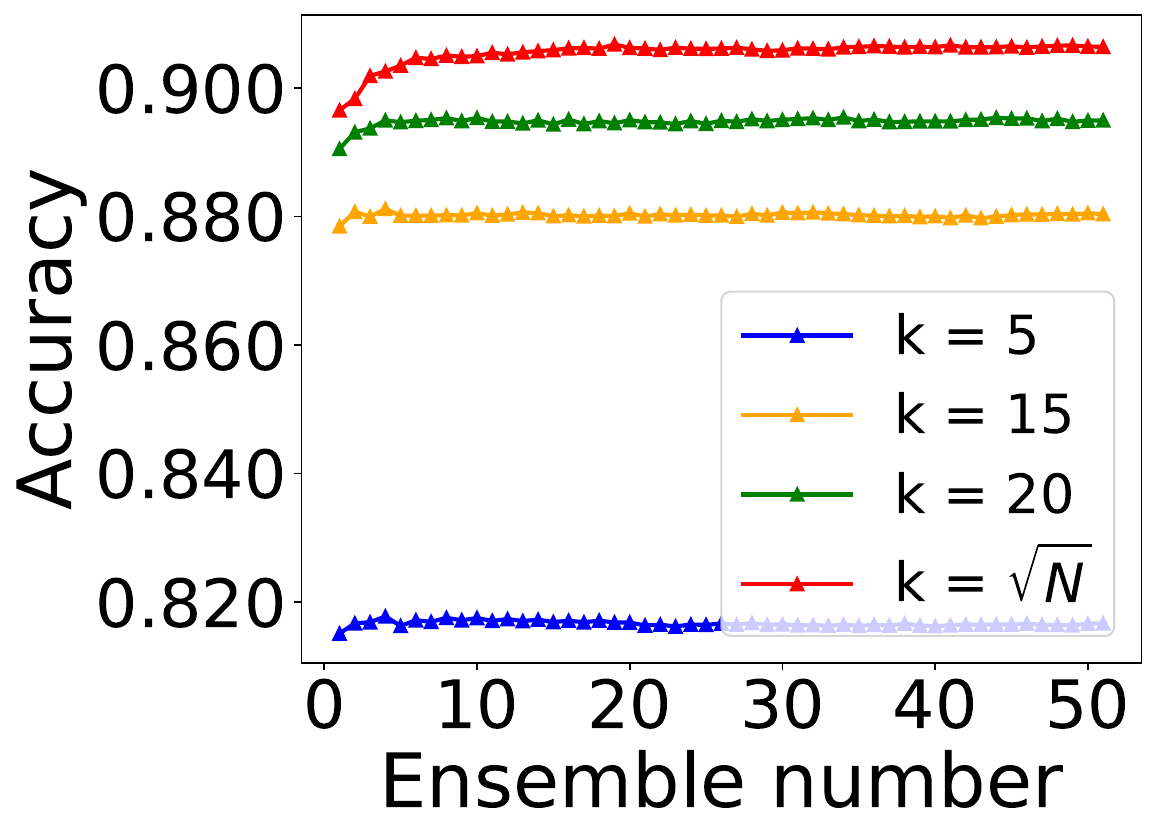}}
\subfloat[LogL Mean K]{\includegraphics[width=0.25\textwidth]{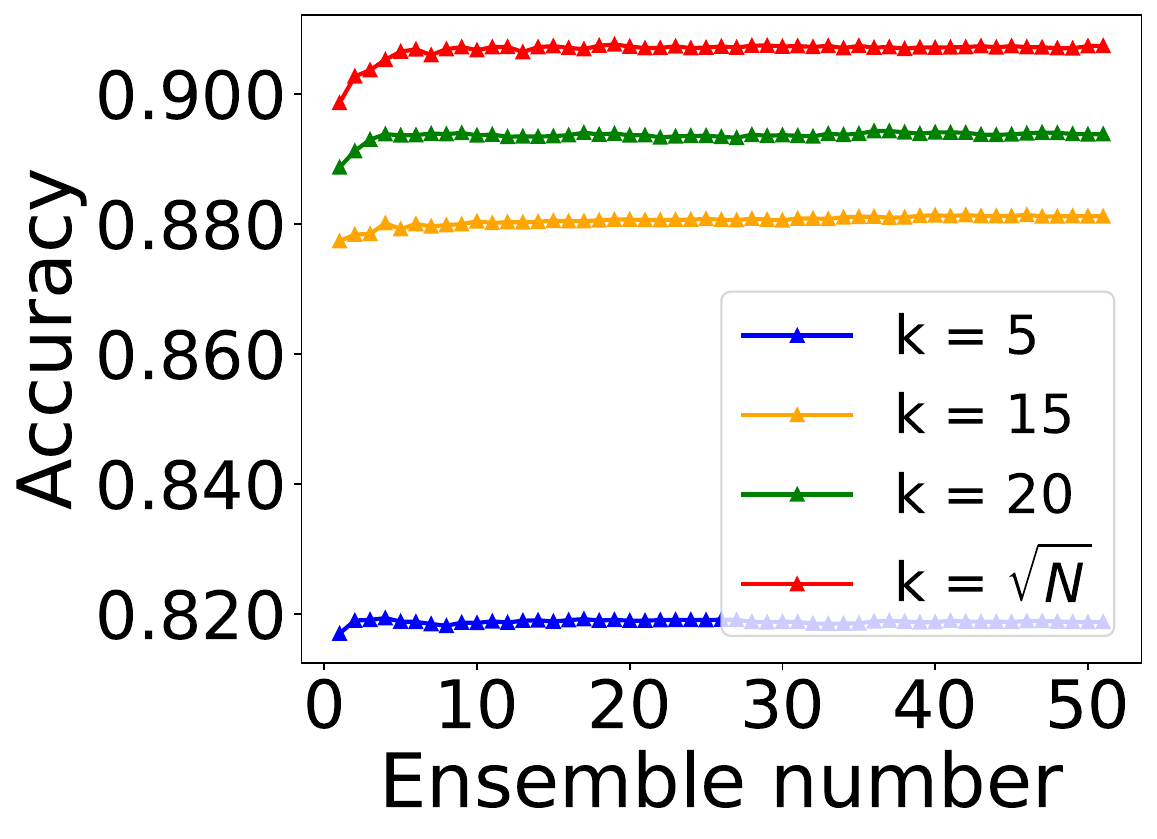}}
\caption{Analysis of K values.}
\label{k_analysis}
\vspace{-20pt}
\end{figure*}

\textbf{Parameter K Analysis.} We show the effect of $K$ on the first data in Figure \ref{k_analysis}. As shown in Figure \ref{k_analysis}, the setting of $K=\lfloor\sqrt{N}\rfloor$ obtains the best performance under the four experimental settings.


\textbf{Comparison with Other Variance Reducing Techniques.} We compare our method with five other widely used regularization techniques, which are Batch-Norm (BN), Droup-Out (DO), BN\&DO, Clip-Grad-Norm (CGN). 
\textbf{BN} is implemented by inserting a normalization layer after each linear transformation. 
\textbf{DO} is applied with a dropout rate set to 0.3.
\textbf{CGN} is applied with a max norm of 1.0.
In the \textbf{BN\&DO} configuration, both Batch Normalization and Dropout are incorporated simultaneously, with their respective parameter settings kept unchanged. 
\textbf{Our} method corresponds to the result when $T=31$ under the exp\_vote\_Accuracy column in table \ref{table_exp_acc}. 
The comparison accuracy results are shown in table \ref{Comparison of methods for reducing variance}, it can be found that our method is significantly superior to the reference methods.

\begin{wraptable}{r}{0.55\textwidth}
\centering
\vspace{-23pt}
    \begin{minipage}{0.55\textwidth}
        \caption{Variance Reduction Comparison Results}
        \label{Comparison of methods for reducing variance}
        \setlength{\tabcolsep}{1.1pt}
        \small
        \renewcommand{\arraystretch}{0.8}
        \begin{tabular}{cccccccc}
            \hline
             &  &Our  &MLP  &BN  &DO  &BN\&DO    & CGN\\
            \hline
            & 1   &\textbf{0.9130}  &0.8910  &0.8793  &0.8907  &0.8946  &0.8944   \\
            & 2   &\textbf{0.8772}  &0.8511  &0.8658  &0.8655  & 0.8646 &0.8688   \\
            & 3   &\textbf{0.6674}  &0.6458  &0.6544  &0.6525  &0.6381  &0.6500   \\
            & 4   &\textbf{0.7873}  &0.7658  &0.7619  &0.7686  &0.7715  &0.7288   \\
            & 5   &\textbf{0.8682}  &0.8557  &0.8557  &0.8541  &0.8560  &0.8526   \\
            & 6   &\textbf{0.9758}  &0.9687  &0.9691  &0.9637  &0.9623  &0.9599   \\
            & 7   &\textbf{0.9797}  &0.9640  &0.8557  &0.9464  &0.9600  &0.9639   \\
            & 8   &\textbf{0.5260}  &0.4850  &0.4842  &0.4807  &0.4895  &0.4860   \\
            & 9   &\textbf{0.8676}  &0.8448  &0.8447  &0.8504  &0.8501  &0.8504   \\
            & 10  &\textbf{0.6148}  &0.5775  &0.5797  &0.5900  &0.5766  &0.5891   \\
            & 11  &\textbf{0.9150}  &0.8448  &0.9023  &0.8970  &0.9096  &0.9035   \\
            & 12  &\textbf{0.9309}  &0.9051  &0.9194  &0.9122  &0.9117  &0.9182   \\
            & 13  &\textbf{0.8801}  &0.8740  &0.8543  &0.8517  &0.8552  &0.8486   \\
            \hline
        \end{tabular}
    \end{minipage}
    \hfill
    \begin{minipage}{0.55\textwidth}
        \captionof{table}{Friedman Statistic Values}
        \label{friedman_test}
        \small
        \setlength{\tabcolsep}{3.5pt}
        \renewcommand{\arraystretch}{0.8}
        \begin{tabular}{ccc}
            \hline
            \textbf{Evaluation metric} & $\boldsymbol{\tau_F}$ & \textbf{Critical value}  \\ 
            \hline
            ExpL Accuracy                     & 28.8              & 2.892                  \\
            ExpL $F_{1}$-score                & 26.2              & 2.892                   \\
            LogL    Accuracy                     & 29.4              & 2.892                   \\
            LogL    $F_{1}$-score                & 24.3              & 2.892                    \\  
            \hline
        \end{tabular}
    \end{minipage}
\vspace{-10pt}
\end{wraptable}

\textbf{Statistical Significance analysis.} 
To evaluate the statistical significance of performance differences between the  RSS-MLP and SRS-MLP, we utilized the Friedman test, a non-parametric method for comparing multiple algorithms across various datasets. The significance is judged by comparing the relationship between the statistic $\tau_F$ and the critical value. We calculate the statistics $\tau_F$ based on Table \ref{table_exp_acc}-Table \ref{table_log_f1}, respectively, and show these values and its corresponding critical values at significance level \( \alpha = 0.05 \) in Table \ref{friedman_test}. As shown in Table \ref{friedman_test}, it is easy to find that in both four settings, the $\tau_F$ values are much higher than the critical values, means that RSS-MLP and SRS-MLP exhibit statistically significant differences in performance.

\begin{figure}[!ht] %
\begin{center}
\includegraphics[width=0.22\textwidth]{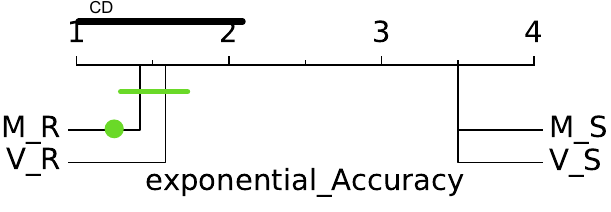}
\includegraphics[width=0.22\textwidth]{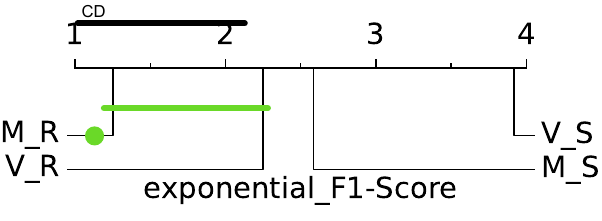}
\includegraphics[width=0.22\textwidth]{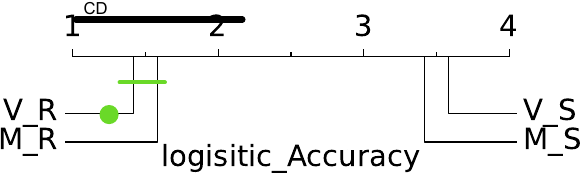}
\includegraphics[width=0.22\textwidth]{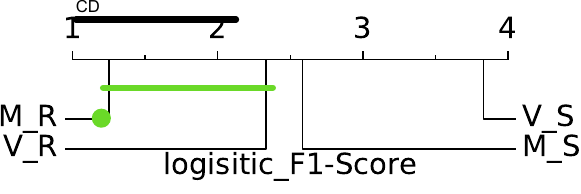}\\
\caption{Critical Difference (CD) Diagram. The abbreviations used in this diagram are: V (Vote): Voting-based method, M (Mean): Mean-based method. S (SRS): simple random sampling, R (RSS): ranked set sampling.}  
\label{cd}
\end{center}
\vspace{-30pt}
\end{figure}

To further analyze these differences, we applied the Nemenyi post-hoc test. If the difference in mean ranks exceeds the critical difference (CD), it indicates a statistically significant performance gap between the models. The Nemenyi post-hoc test results are shown in Figure \ref{cd}. In Figure \ref{cd}, the horizontal line at the top of each subgraph denotes the critical difference (CD). Algorithms that are marked with a green star are identified as the best-performing ones. If an algorithm is not connected to the green star by a green line, it indicates that there is a significant difference in performance compared to the best algorithm. The findings reveal that the average ranking of the RSS-MLP method consistently surpasses that of the SRS-MLP method, and in most cases, its performance is significantly better than that of the voting strategy.

\section{Conclusion}


In this paper, a generalization error bound is established, revealing the impact of loss variance under empirical distribution on generalization performance. Guided by this bound, RSS sampling is utilized to reduce the variance of the multilayer perceptron. Theoretically, the variance of empirical convex loss with RSS sampling is proven to be smaller than that with SRS sampling. Then a RSS-MLP method is proposed that integrates multiple MLPs trained on RSS sampling data sets. Through extensive experiments on multiple datasets, the effectiveness and rationality of RSS-MLP are demonstrated. Future work will focus on continuing to enhance the generalization ability of MLP models from the perspective of variance reduction.

\section{Appendix}

\subsection{Proof of Theorem 1}\label{proof-of-theorem-1}

In addition to the use of bounded difference property of  McDiarmid inequality \cite{Devroye1996APT}, we further require the loss function is classification-calibrated and additive. These two properties are stated as follows.

\begin{definition}(\textbf{Classification-Calibrated}~\cite{Bartlett2006ConvexityCA})
A loss function $\phi(f(\boldsymbol{\mathit{X}}), Y)$ is called classification-calibrated if minimizing the expected risk results in a classifier:
\begin{equation}
\hat{f} = \arg \min_f \mathbb{E}[\phi(f(\boldsymbol{\mathit{X}}  ), Y)],
\end{equation}
that matches the optimal decision boundary: $\hat{Y} = \text{sign}(\hat{f}(\boldsymbol{\mathit{X}}))= \text{sign}(2\eta(\boldsymbol{\mathit{X}}  ) - 1)$, where $\eta(X) =\mathbb{P}(Y = 1 |\boldsymbol{\mathit{X}}  = \boldsymbol{\mathit{x}})$ is the posterior probability.
\end{definition}

\begin{definition}(\textbf{Additive}~\cite{Mohri2018VC})
When the risk function $\phi(f)$ is additive means that the empirical loss value on a sample with
$N$ objects can be calculated by the sum of the losses on each object: $\hat{\mathit{R}}_{\phi}(f)=\frac{1}{N} \sum_{i=1}^{N} \phi(y_i,f(\textit{\textbf{x}}_i))$.

\end{definition}

Let the expected convex risk be $ \mathit{R}_{\phi}(f) = \mathbb{E}_{\phi}(Yf(X)) $. If a risk function $\phi(f)$ is additive and the samples are independent identically distributed, then the empirical value of $\phi(f)$ is an unbiased estimator. That is, it satisfies:
{\small
\begin{align}
\label{unbiasphi}
\mathbb{E}\hat{\mathit{R}}_{\phi}(f)
= \mathbb{E}\frac{1}{N}\sum_{i=1}^{N}\phi(y_{i},f(\textit{\textbf{x}}_{i}))
= \frac{1}{N} \sum_{i=1}^{N} \mathbb{E}( \phi(y_{i},f(\textit{\textbf{x}}_{i}))) 
=\mathbb{E}( \phi(y,f(\textit{\textbf{x}})))=\mathit{R}_{\phi}(f) .
\end{align}
}
\noindent\textbf{Theorem \ref{th1}.}
\emph{Let \( \boldsymbol{\mathit{X}}_1, \boldsymbol{\mathit{X}}_2, \ldots, \boldsymbol{\mathit{X}}_N \) be independent identically distributed random variables. Let $\mathcal{F}: \mathcal{X}^d \rightarrow \mathbb{R}$ denote the hypothesis space containing the hypothesis function $f(\textit{\textbf{x}})$ with a bounded norm. Assume that $\phi(\textit{\textbf{x}})$ is an additive loss function, and it is classification-calibrated
and Lipschitz continuous with a constant $L$. Then, with probability at least $1 - \delta$ (where $\delta > 0$), we have:
{\small
    \begin{align}
        \notag\mathit{R}(f) - \mathit{R}^{*}
        \le \psi^{-1} \bigg (2\underset{f\in \mathcal{F}}{\sup}\sqrt{ \frac{ 2\left(L\max_\textit{\textbf{x}}\left\|f(\textit{\textbf{x}}) \right\| \right)^2}{N}
        \ln{\frac{1}{\delta }}}
        +2\underset{f\in \mathcal{F}}{\sup}\sqrt{ \mathbb{V} \left ( \hat{\mathit{R}}_{\phi}(f) \right ) }  +\inf_{f\in \mathcal{F}} \mathit{R}_{\phi}(f) - \mathit{R}_{\phi}^{*} \bigg),
    \end{align}
}    
where $ \mathbb{V} (\cdot)$ is the variance, $\psi$ is a nondecreasing function, $ \mathit{R}_{\phi}(f) = \mathbb{E}_{\phi}(Yf(X)) $ is the expected convex risk, and
$\mathit{R}_{\phi}^{*}= \inf \mathit{R}_{\phi}(f)$ is the global optimal convex risk.
}

\begin{proof}
According to the Theorem \ref{th1} in ~\cite{Bartlett2006ConvexityCA}, we can convert the convergence in the sense of zero-one loss to that of $\phi$ loss according to a non-decreasing function $\psi$:
    \begin{equation}
    \label{convexphi}
        \mathit{R}(f) - \mathit{R}^{*}
        \le\psi^{-1}\left ( \mathit{R}_{\phi}(f)-\mathit{R}_{\phi}^{*}\right )
    \end{equation}

Usually, the excess risk can be divided into the estimation error and approximation error~\cite{Devroye1996APT}:
\begin{align}
\label{r1}
 & \mathit{R}_{\phi}({f})-\mathit{R}_{\phi}^{*} 
 =  \underset{Estimation\ error}{\underbrace{\mathit{R}_{\phi}({f})- \underset{f\in \mathcal{F}}{\inf}\mathit{R}_{\phi}(f)}}+\underset{Approximation\ error}{\underbrace{\underset{f\in \mathcal{F}}{\inf}\mathit{R}_{\phi}(f) - \mathit{R}_{\phi}^{*}}}.
\end{align}

Because of the approximation error maybe not controllable, so we generally consider to control the estimation error. By Lemma 8.2 in \cite{Devroye1996APT}, we further have:
\begin{align}
\label{r2}
\mathit{R}_{\phi}({f})- \underset{f\in \mathcal{F}}{\inf}\ \mathit{R}_{\phi}(f)\le 2\underset{f\in \mathcal{F}}{\sup}\left | \mathit{R}_{\phi}(f)-\hat{\mathit{R}}_{\phi}(f) \right |.
\end{align}

By Eq. (\ref{unbiasphi}), we subtract and add one item at a time, then we have:
{\fontsize{8pt}{8pt}\selectfont
\begin{align}
\label{x-ex+ey}
\left|\mathit{R}_{\phi}(f)-\hat{\mathit{R}}_{\phi}(f) \right |  
\le \left|\hat{\mathit{R}}_{\phi}(f)-\mathit{R}_{\phi}(f)\right| -\mathbb{E}\left (\left|\hat{\mathit{R}}_{\phi}(f)-\mathit{R}_{\phi}(f)\right|\right) 
+\mathbb{E}\left(\left|\hat{\mathit{R}}_{\phi}(f)-\mathbb{E}\ \hat{\mathit{R}}_{\phi}(f) \right|\right).
\end{align}
}

To bound the first two terms of Eq. (\ref{x-ex+ey}), according to McDiarmid inequality, we need to give the difference bound. Let $D' =\{(\textbf{\textit{x}}_1, y_1),\ldots, (\textbf{\textit{x}}_i', y_i'),\ldots,(\textbf{\textit{x}}_N, y_N)\}$ be the different set
and $\hat{\mathit{R}}_{\phi}'(f)$ be the empirical loss estimated by it, then we have:

\begin{align}
\label{ci}
\left||\hat{\mathit{R}}_{\phi}(f)-\mathbb{E}\ \hat{\mathit{R}}_{\phi}(f)|-|\hat{\mathit{R}}_{\phi}'(f)-\mathbb{E}\ \hat{\mathit{R}}_{\phi}'(f)|\right|  
&\le  \left|\hat{\mathit{R}}_{\phi}(f)-\hat{\mathit{R}}_{\phi}'(f)\right|
 \le\frac{ L}{N}\left\|f(\textit{\textbf{x}})-f (\textit{\textbf{x}}')\right\|  \\
&\le \frac{ L}{N} \left(\left\|f(\textit{\textbf{x}})\right\| +\left\|f(\textit{\textbf{x}}')\right\|\right)
\le \frac{ 2L}{N}\max_{\textit{\textbf{x}}}\left\|f(\textit{\textbf{x}})\right\|,\notag
\end{align}
where the first inequality and the third one are according to the triangle inequality, and the second is according to the L-Lipschitz continuity.

Then according to McDiarmid inequality, for $\delta> 0$, with a probability at least $1-\delta$ over the i.i.d. sample set, we have:
\begin{align}
\label{x-ex}
\left|\hat{\mathit{R}}_{\phi}(f)-\mathit{R}_{\phi}(f)\right| -\mathbb{E}\left (\left|\hat{\mathit{R}}_{\phi}(f)-\mathit{R}_{\phi}(f)\right|\right) 
\le\sqrt{ \frac{2\left(L\max_\textit{\textbf{x}}\left\|f(\textit{\textbf{x}}) \right\| \right)^2 }{N} \ln{\frac{1}{\delta }}}.
\end{align}

To bound the third term of Eq. (\ref{x-ex+ey}), we use the Jensen's inequality,
\begin{align}
\label{r4}
\mathbb{E}\left(\left|\hat{\mathit{R}}_{\phi}(f)-\mathbb {E}\ \hat{\mathit{R}}_{\phi}(f) \right|\right) 
\le  \sqrt{\mathbb{E}\left(\left|\hat{\mathit{R}}_{\phi}(f)-\mathbb {E}\ \hat{\mathit{R}}_{\phi}(f) \right|\right)^2 }
=  \sqrt{\mathbb {V} \left ( \hat{\mathit{R}}_{\phi}(f) \right ) }.
\end{align}
Combining Eq. (\ref{convexphi})-Eq. (\ref{r4}), we obtain the final result.
\end{proof}

\subsection{Proof of Theorem 2}\label{proof-of-theorem-2}

\noindent\textbf{Theorem 2. }(\textbf{Equal Expectation})
\emph{For the additive convex loss function $\phi$ and an independent identically distribution sample, the expectations of empirical risks based on the SRS sample and RSS sample are equal: $\mathbb{E} \hat{R} _{\phi,RSS}(f) = \mathbb {E}\hat{R}_{\phi,SRS}(f)$ .
}

\begin{proof}
For the objects with the identity distribution and the loss with additive property, from Eq. (\ref{unbiasphi}), it is known that: $\mathbb{E}\hat{R }_{\phi,SRS}(f)=\mathbb{E}( \phi(y,f(\textit{\textbf{x}})))$.

Let the joint probability distribution of sample pair $(X,Y)$ be $g(\textit{\textbf{x}},y)$
and the joint probability distribution of r-th order statistic of $(X,Y)$ be $g_{[r]}(\textit{\textbf{x}},y)$, we have~\cite{Chen1999DensityEU}:
\begin{equation}
g(\textit{\textbf{x}},y)  = \frac{1}{K}\sum_{r=1}^{K}g_{[r]}(\textit{\textbf{x}},y),
\label{PDF}
\end{equation}
where $1\leq r\leq K$.
Thus, for the RSS sample, we have
\begin{align}
\label{expec}
\mathbb{E}\hat{R}_{\phi,RSS}(f)
&= \mathbb{E}\frac{1}{K}\frac{1}{m}\sum_{r=1}^{K}\sum_{i=1}^{m} \phi(y_{[r]i},f(\textit{\textbf{x}}_{[r]i}))
=\frac{1}{K} \sum_{r=1}^{K} \mathbb{E}\phi(y_{[r]},f(\textit{\textbf{x}}_{[r]}))\\ \notag
&=\frac{1}{K} \sum_{r=1}^{K} \int \phi(y,f(\textit{\textbf{x}})) g_{[r]}(f(\textit{\textbf{x}}),y) d\textit{\textbf{x}}
= \int \phi(y,f(\textit{\textbf{x}})) g (f(\textit{\textbf{x}}),y) d\textit{\textbf{x}}\\\notag
&=\mathbb{E}( \phi(y,f(\textit{\textbf{x}}))),\notag
\end{align}
where $m$ is the sample size of r-th order statistic and $m=\lfloor N/K \rfloor$.
\end{proof}

\subsection{Proof of Theorem 3}\label{proof-of-theorem-3}

\noindent\textbf{Theorem 3. }(\textbf{Smaller Variance})
\emph{
For the additive convex loss function $\phi$ and an independent identically distribution sample, the variances of the empirical risk based on the SRS and RSS samples satisfy: $\mathbb{V} \hat{R}_{\phi,RSS}(f) \le \mathbb{V} \hat{R}_{\phi,SRS}(f)$.
}

\begin{proof}
Let $\mathcal{G}$ be an universal reproducing kernel Hilbert space,
the independent of sample~\cite{Gretton2007} means that if the object $\textbf{\textit{x}}_i$ and $\textbf{\textit{x}}_j$ are independent, then for any $g \in  \mathcal{G}$,  $g(\textbf{\textit{x}}_i)$ and $g(\textbf{\textit{x}}_j)$ are independent. Thus, we have,
{\fontsize{7pt}{7pt}\selectfont
\begin{align}\label{Var_srs}
     \mathbb{V}  \hat{R }_{\phi,SRS}(f)
    =   \mathbb{V}  \left( \frac{1}{N}\sum_{i=1}^{N} \ell(y_{i},f(\textit{\textbf{x}}_{i}))\right ) 
    =  \frac{1}{N}\mathbb{V} \left( \ell(y,f(\textit{\textbf{x}}))\right) 
     =  \frac{1}{N} \left\{\mathbb{E}\left(\phi\left(y,f(\textit{\textbf{x}})\right)\right)^2 - \left[\mathbb{E}  \phi\left(y,f(\textit{\textbf{x}})\right)\right]^2 \right\},
\end{align}
}
where the second inequality is according to the independent of samples. It is worth mentioning that although the function $f$ maybe learnt by the $ N$ objects, under the independent assumption of sample, the second inequality is satisfied.
For the RSS sample, the variance is:
\begin{align} \label{Var_rss}
&\mathbb{V}  \hat{R}_{\phi,RSS}(f) = \frac{1}{K^2m} \sum_{r=1}^{K} \mathbb{V} \big( \phi(y_{[r]},f(\textit{\textbf{x}}_{[r]}))\big) \\ \notag
&= \frac{1}{K^2m} \sum_{r=1}^{K} \left\{ \mathbb{E}\left(\phi(y_{[r]},f(\textit{\textbf{x}}_{[r]}))\right)^2 - \left[\mathbb{E}(\phi(y_{[r]},f(\textit{\textbf{x}}_{[r]})))\right]^2 \right\}  \\ \notag
&=\frac{1}{Km} \left\{  \mathbb{E}\left(\phi(y,f(\textit{\textbf{x}}))\right)^2-\frac{1}{K} \sum_{r=1}^{K} \left[\mathbb{E}\left(\phi(y_{[r]},f(\textit{\textbf{x}}_{[r]})\right)\right]^2 \right\} \\ \notag
&= \mathbb{V}  \hat{R }_{\phi,SRS}(f)+ \frac{1}{Km} \left\{ \left[\mathbb{E} \phi(y,f(\textit{\textbf{x}}))\right]^2 - \frac{1}{K}\sum_{r=1}^{K}\left[\mathbb{E}(\phi(y_{[r]},f(\textit{\textbf{x}}_{[r]}))) \right] ^2 \right\} ,
\end{align}
where the third equality is according to Eq. (\ref{PDF}).

Further, by Eq. (\ref{PDF}) and Jensen inequality, we have:
\begin{align}
  \left [ \mathbb{E}\phi(y,f(\textit{\textbf{x}}))\right ]^{2}
=  \left [ \frac{1}{K} \sum_{r=1}^{K} \mathbb{E}\phi(y_{[r]},f(\textit{\textbf{x}}_{[r]})) \right ]^{2}
  \leq  \frac{1}{K} \sum_{i=1}^{K}  \left [\mathbb{E}  \phi(y_{[r]},f(\textit{\textbf{x}}_{[r]})) \right ]^{2}.
\end{align}
Then, we have $\mathbb{V}   \hat{R}_{\phi,RSS}(f)\leq \mathbb{V}  \hat{R }_{\phi,SRS}(f)$.
\end{proof}

Before proving Examples 1 and Examples 2, we show well-know Taylor expansion.
The Taylor expansion of the function $h(u)$ at $u = 0$ can be expressed as:
\begin{align}
\label{Taylor}
h(u) = \sum_{s=0}^{n}\frac{h^{s}(0)}{(s)!}u^{s} +O(u^{n+1})
= \underset{a_{h}}{\underbrace{\sum_{s=0}^{\infty}\frac{h^{2s}(0)}{(2s)!}}}u^{2s}+\underset{b_{h}}{\underbrace{\sum_{s=1}^{\infty}\frac{h^{2s+1}(0)}{(2s+1)!}}}u^{2s+1}.
\end{align}
Then, we use the well-know Taylor expansion to observe the representation of the function $\left[\mathbb{E} \phi(y,f(\textit{\textbf{x}}))\right]^2 - \frac{1}{K}\sum_{r=1}^{K}\left[\mathbb{E}(\phi(y_{[r]},f(\textit{\textbf{x}}_{[r]}))) \right] ^2$.

Let $z=-yf(\textit{\textbf{x}})$ and $z_{[r]}=-y_{[r]}f(\textit{\textbf{x}}_{[r]})$. It is known that $z^2=1$ and $z_{[r]}^2=1$ when $y, f(\textit{\textbf{x}})\in\{-1,+1\}$ in the binary classification sense. By Eq. (\ref{Taylor}),
for any convex loss function, we can rewrite the given expression $\left[\mathbb{E} \phi(y,f(\textit{\textbf{x}}))\right]^2 - \frac{1}{K}\sum_{r=1}^{K}\left[\mathbb{E}(\phi(y_{[r]},f(\textit{\textbf{x}}_{[r]}))) \right] ^2$ as follows:
{\fontsize{8pt}{8pt}\selectfont
\begin{align}
\label{Taylor_E}
&\left[\mathbb{E} \phi(y,f(\textit{\textbf{x}}))\right]^2 - \frac{1}{K}\sum_{r=1}^{K}\left[\mathbb{E}(\phi(y_{[r]},f(\textit{\textbf{x}}_{[r]}))) \right] ^2=\left[\mathbb{E} (a_{\phi}+b_{\phi}z)\right]^2- \frac{1}{K}\sum_{r=1}^{K}\left[\mathbb{E}(a_{\phi}+b_{\phi}z_{[r]}) \right] ^2\\ \notag
=&(a_{\phi}+b_{\phi}\mathbb{E}z)^2- \frac{1}{K}\sum_{r=1}^{K}(a_{\phi}+b_{\phi}\mathbb{E}z_{[r]})^2\\ \notag
=&(a_{\phi}^2+2a_{\phi}b_{\phi}\mathbb{E}z+b_{\phi}^2(\mathbb{E}z)^2)- \frac{1}{K}\sum_{r=1}^{K}(a_{\phi}^2+2a_{\phi}b_{\phi}\mathbb{E}z_{[r]}+b_{\phi}^2(\mathbb{E}z_{[r]})^2)\\ 
=&b_{\phi}^2\left ( (\mathbb{E}z)^2- \frac{1}{K}\sum_{r=1}^{K}(\mathbb{E}z_{[r]})^2\right ).\notag
\end{align}
}
Based on the above equation, we can prove the following specific formulations.

\subsection{Proof of Example 1.}\label{proof-of-examples-1}

 (\textbf{Example 1. Exponential Loss})
For the  exponential loss function $\phi (\alpha)= e^{- \alpha}$ with the property of convexity and additive capability, Exponential loss is globally $L$-Lipschitz continuous with $L = C\exp(RC)$, where  $C=\max_i \| \textit{\textbf{x}}_i \|$ (data-dependent) and $R=\max_f \| f(\textit{\textbf{x}}) \|$. According to Theorem 1, the variance of empirical risk estimation $\mathbb{n}[\hat{R}_{\phi, \mathrm{RSS}}]$ and $\mathbb{n}[\hat{R}_{\phi, \mathrm{SRS}}]$ reflects the model's generalization ability. Therefore, comparing these two variances allows for an evaluation of the differences in generalization performance across various sampling strategies. we have,
{\fontsize{7pt}{7pt}\selectfont
    \begin{equation}
    \label{Exp_Var}
        \begin{split}
          \mathbb{V} \hat{R }_{\phi,RSS}(f)
          &=  \mathbb{V} \hat{R }_{\phi,SRS}(f)+\left(\frac{e-e^{-1}}{2}\right)^{2}\frac{1}{K}\frac{1}{m}\left [ \left [\mathbb{E}(yf(\textit{\textbf{x}}))\right ]^{2}-\frac{1}{K} \sum_{r=1}^{K} \left [\mathbb{E}(y_{[r]}f(\textit{\textbf{x}}_{[r]}))\right ]^{2} \right ].
        \end{split}
    \end{equation}
}

\begin{proof}
In this proof, we will use the exponential loss function as a specific example of a convex loss function to illustrate the argument.
  \begin{align}
    \label{Exp_SRS_R}
    \hat{R }_{\phi,SRS}(f) &= \frac{1}{N} \sum_{i=1}^{N}e^{- y_if(\textit{\textbf{x}}_{i})},\\ \notag
    \hat{R }_{\phi,RSS}(f) &= \frac{1}{K}\frac{1}{m}\sum_{r=1}^{K}\sum_{i=1}^{m} e^{ -y_{[r]i}f(\textit{\textbf{x}}_{[r]i})}.
  \end{align}

According to Eq. (\ref{Var_rss}), we have:
\begin{align} \label{Exp_RSS_Var}
&\mathbb{V}  \hat{R }_{\phi,RSS}(f)
=  \mathbb{V}  \hat{R }_{\phi,SRS}(\textit{\textbf{x}})
+ \frac{1}{K}\frac{1}{m}\left\{\left[\mathbb{E} e^{-yf(\textit{\textbf{x}})}\right]^2 - \frac{1}{K}\sum_{r=1}^{K}\left[\mathbb{E}\left(e^{- y_{[r]} f(\textit{\textbf{x}}_{[r]})}\right)\right]^2 \right\}.
\end{align}

When the loss function is specifically an exponential loss, by some simple calculation, we have $b_{\phi} = \frac{e-e^{-1}}{2}$ in Eq. (\ref{Taylor}). Then, the above equation can be rewritten as follows:

{\fontsize{7pt}{7pt}\selectfont
\begin{align}
\label{q1}
\left[\mathbb{E} e^{-yf(\textit{\textbf{x}})}\right]^2 - \frac{1}{K}\sum_{r=1}^{K}\left[\mathbb{E}\left(e^{- y_{[r]} f(\textit{\textbf{x}}_{[r]})}\right)\right]^2
= \bigg(\frac{e-e^{-1}}{2}\bigg)^2\left [\left(\mathbb{E}(yf(\textit{\textbf{x}}))\right)^2-\frac{1}{K} \sum_{r=1}^{K} \left(\mathbb{E}(y_{[r]}f(\textit{\textbf{x}}_{[r]}))\right)^2\right ]
\end{align}
}

Finally, combining Eq. (\ref{Exp_RSS_Var}) and Eq. (\ref{q1}), we obtain the final result.

\end{proof}

\subsection{ Proof of Example 2.}
\label{proof-of-examples-2}

\textbf{Example 2. (Logistic Loss)}
For the logistic loss function $\phi (\alpha)= log(1+e^{- \alpha})$ with the property of convexity and additive capability, Logistic loss is globally $L$-Lipschitz continuous with $L = C$, where  $C=\max_i \| \textit{\textbf{x}}_i \|$ (data-dependent). According to Theorem 1, the variance of empirical risk estimation $\mathbb{n}[\hat{R}_{\phi, \mathrm{RSS}}]$ and $\mathbb{n}[\hat{R}_{\phi, \mathrm{SRS}}]$ reflects the model's generalization ability. Therefore, comparing these two variances allows for an evaluation of the differences in generalization performance across various sampling strategies. we have,
    \begin{equation}
    \label{Log_Var}
        \begin{split}
          &\mathbb{V} \hat{R }_{\phi,RSS}(f)
          =  \mathbb{V} \hat{R }_{\phi,SRS}(f) +\frac{1}{4Km}\left [\left(\mathbb{E}(yf(\textit{\textbf{x}}))\right)^{2}-\frac{1}{K} \sum_{r=1}^{K} \left(\mathbb{E}(y_{[r]}f(\textit{\textbf{x}}_{[r]}))\right)^{2}\right ].
        \end{split}
    \end{equation}

\begin{proof}
In this proof, we will use the logistic loss function as a specific example of a convex loss function to illustrate the argument.
    \begin{align}
        \label{Log_SRS_R}
        \hat{R }_{\phi,SRS}(f) &= \frac{1}{N} \sum_{i=1}^{N}log(1+e^{- y_if(\textit{\textbf{x}}_{i})}),\\
        \hat{R }_{\phi,RSS}(f) &= \frac{1}{K}\frac{1}{m}\sum_{r=1}^{K}\sum_{i=1}^{m} log(1+e^{ -y_{[r]i}f(\textit{\textbf{x}}_{[r]i})}).
    \end{align}
According to Eq. (\ref{Var_rss}), we have:
{\fontsize{6pt}{6pt}\selectfont
\begin{align} \label{Log_RSS_Var}
\mathbb{V}  \hat{R }_{\phi,RSS}(f)
&=  \mathbb{V}  \hat{R }_{\phi,SRS}(f)+\frac{1}{K}\frac{1}{m}\bigg \{
\left[\mathbb{E}log(1+e^{ -yf(\textit{\textbf{x}})}) \right]^2 
-\frac{1}{K}\sum_{r=1}^{K}\left[\mathbb{E}\left(log(1+e^{ -y_{[r]}f(\textit{\textbf{x}}_{[r]})})\right)
\right]^2 \bigg \}.
\end{align}
}
The Taylor expansion of the $log(1+e^{-u})$ function is expressed as:
\begin{eqnarray}
log(1+e^{-u})
 = & \left\{\begin{matrix}
  & -\frac{EulerE[n-1,0]}{2n(n-1)!}u^{n} &n\ge1 \\
  & log2  &n=0,
\end{matrix}\right.
\label{Taylor_L}
\end{eqnarray}
where EulerE is the Euler function. When $n$ is an odd number larger than $3$, $EulerE[n-1,0]=0$.
Thus, we get $b_{\phi} = -\frac{1}{2}$. Then, Eq. (\ref{Taylor_E}) can be rewritten as follows:
\begin{align}
\label{log+E}
&\left[\mathbb{E}log(1+e^{ -yf(\textit{\textbf{x}})}) \right]^2 - \frac{1}{K}\sum_{r=1}^{K}\left[\mathbb{E}\left(log(1+e^{ -y_{[r]}f(\textit{\textbf{x}}_{[r]})})\right)\right]^2 \\\notag
&=\left(- \frac{1}{2}\right)^2\left [ \left [\mathbb{E}(yf(\textit{\textbf{x}}))\right ]^{2}-\frac{1}{K} \sum_{r=1}^{K} \left [\mathbb{E}(y_{[r]}f(\textit{\textbf{x}}_{[r]}))\right ]^{2} \right ]
\end{align}
Finally, Eq. (\ref{Log_RSS_Var}) and Eq. (\ref{log+E}), we obtain the final result.

\end{proof}

\bibliography{rcc}


\end{document}